\documentclass[runningheads]{llncs}

\PassOptionsToPackage{table}{xcolor}
\usepackage{eccv}

\usepackage{eccvabbrv}

\usepackage{graphicx}
\usepackage{booktabs}
\usepackage{multirow}
\usepackage{textcomp}
\usepackage{gensymb}
\usepackage{longtable}
\usepackage{algorithm}
\usepackage{algpseudocode}

\usepackage[accsupp]{axessibility}  %
\usepackage{enumitem}

\usepackage{pgffor}
\usepackage{array}
\usepackage{tabularx}
\usepackage{wrapfig,lipsum}
\usepackage{bbm}
\newcolumntype{Y}{>{\centering\arraybackslash}X}
\newcolumntype{x}[1]{>{\centering\arraybackslash\hspace{0pt}}p{#1}}
\newcolumntype{P}[1]{>{\centering\arraybackslash}p{#1}}

\usepackage[pagebackref,breaklinks,colorlinks,citecolor=eccvblue]{hyperref}

\usepackage{orcidlink}
\usepackage[export]{adjustbox} %

\definecolor{condcolor}{RGB}{230,140,20}  %
\definecolor{gtcolor}{RGB}{34,139,80}     %
\definecolor{predcolor}{RGB}{200,50,50}   %

\newif\ifshowedits
\showeditstrue

\newcommand{\coolname}{\textit{E³C}}

\begin{document}

\title{\coolname: Video Generation with 3D Environmental Memory and Ego-Exo Human Pose Control}

\titlerunning{\coolname: Video Generation with 3D Environmental Memory}

\author{Qiao Gu\inst{1,2}\thanks{Work done during internship at Reality Labs, Meta.} \and  Lingni Ma \inst{1} \and Adam W Harley \inst{1} \and Richard Newcombe \inst{1} \and \\ Florian Shkurti \inst{2} \and Julian Straub \inst{1}}

\authorrunning{Q.~Gu et al.}

\institute{Meta Reality Labs, Redmond, WA 98052, USA\\ \and University of Toronto, Toronto, ON M5S 1A1, Canada \\
\email{q.gu@mail.utoronto.ca, \{lingni.ma, aharley, newcombe\}@meta.com, florian@cs.toronto.edu, jstraub@meta.com}}

\maketitle

\begin{figure}[h!]
\centering
\vspace{-1.5em}
\includegraphics[width=\textwidth,trim={0in 8.3in 0 0},clip]{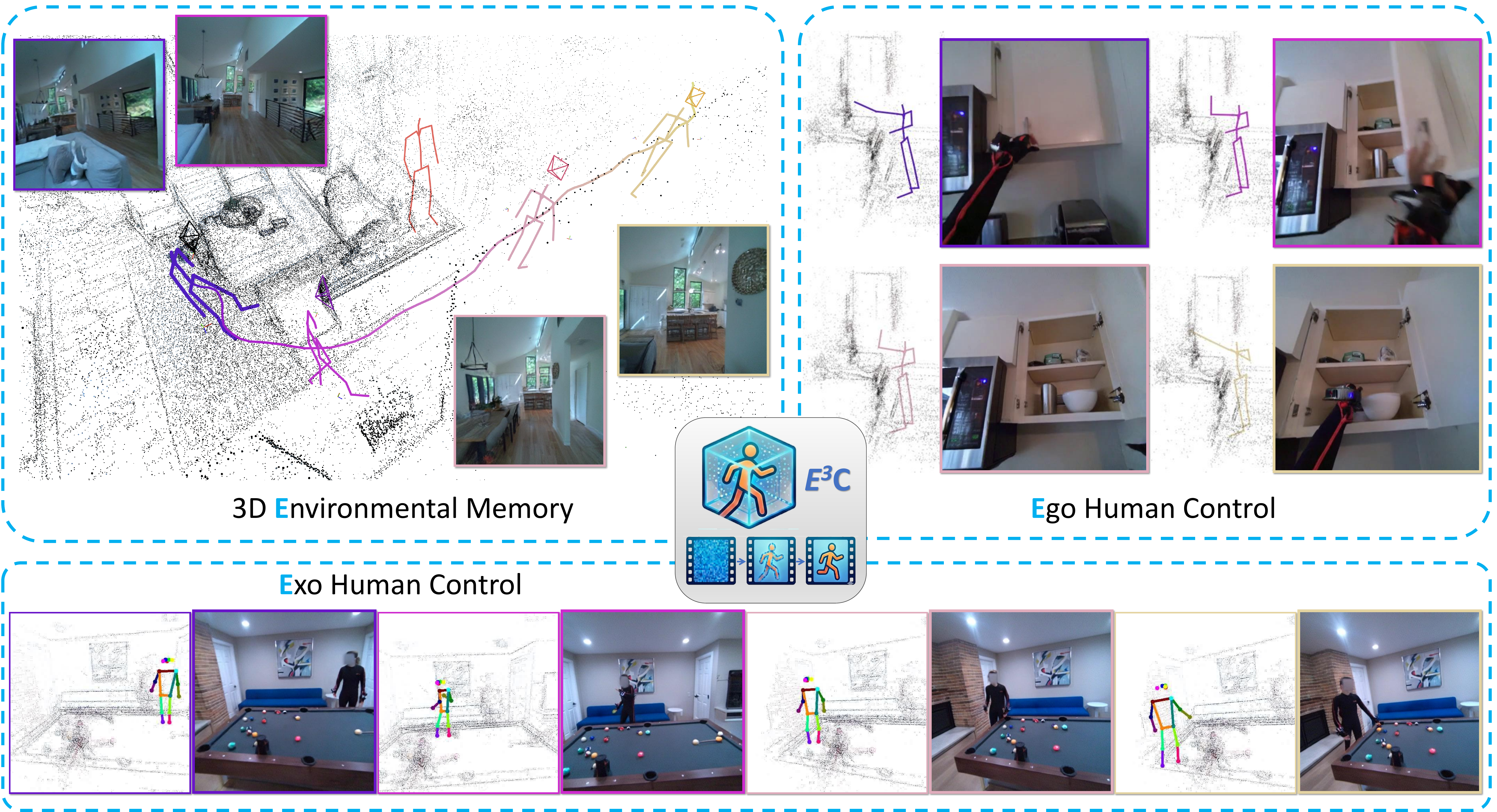}
\vspace{-2em}
\caption{We present \textbf{\coolname{}}, a controllable egocentric video generation model that unifies 3D \textbf{E}nvironmental memory with \textbf{E}go--\textbf{E}xo human pose \textbf{C}ontrol. \coolname{} generates egocentric videos that maintain 3D scene consistency (top left), follow ego-body motion (top right), and adhere to exo-human motion (bottom). All images are from generated videos. Project page: \href{https://e3c-videogen.github.io}{e3c-videogen.github.io}.
\vspace{-3.5em}}
\label{fig:overview}
\end{figure}

\begin{abstract}

Controllable and physically grounded egocentric video generation is essential for embodied agents to reason about how their own and others’ actions manifest and change the world.
Compared to generic video synthesis, egocentric generation is especially challenging:
the camera is tightly coupled to the actor, leading to rapid viewpoint changes and frequent self-occlusions;
the underlying actions are subtle, articulated, and often only partially visible; and both the people and the scene state must evolve consistently with the specified controls.
We present \coolname{}, a controllable video diffusion framework for egocentric generation that builds structured and compact conditions disentangling persistent scene structure from human-driven dynamics.
From context frames, \coolname{} constructs a semi-dense point cloud-based 3D memory and augments each point with appearance descriptors from video-VAE features.
Rendering this memory into target viewpoints produces conditioning aligned with the target frames.
Human dynamics are modeled separately. The observed people in the scene are controlled by skeleton renderings (exo human control), while the camera wearer is specified by their 3D body joints and 6DoF wrist motion (ego human control).
To preserve ego human control when the wearer's body parts are invisible, we introduce an ego motion encoder that produces persistent cross-attention tokens.
Experiments on Nymeria show that \coolname{} improves visual fidelity, camera-motion accuracy, object consistency, and ego \& exo human control over strong baselines, while also enabling intuitive scene editing.
\vspace{-0.5em}
\keywords{Egocentric Vision \and Video Generation}
\vspace{-0.5em}
\end{abstract}

\section{Introduction}
\label{sec:intro}

Embodied agents benefit from being able to imagine how candidate actions will change future first-person observations~\cite{finn2017dvf, koh2021pathdreamer}. A controllable and physically grounded egocentric video generator can serve as a rollout simulator~\cite{yu2025egosim, bai2025peva, tu2025playerone}: given candidate future human controls from a planner, a user, or a motion prior, it visualizes the resulting first-person video for the camera wearer, nearby people, and the surrounding scene. This capability matters in embodied planning, human-robot interaction, AR/VR, and controlled data generation, where action and perception are tightly coupled~\cite{yang2025egolife, fang2024egopat3dv2, escobar2025egocast}.

Egocentric video generation is especially challenging because the camera is physically coupled to the actor. Every reach, turn, and step changes not only the body's configuration but also the viewpoint itself, yielding rapid camera motion, frequent self-occlusion, and large appearance changes across time. At the same time, the underlying actions are complex, subtle, and highly articulated, and their visual consequences may be occluded, delayed, or occur partially or entirely outside the camera’s field of view.
To be useful, a first-person video generator must also make both the humans and the environment respond coherently to the input controls. If the wearer turns, reaches, or walks past another person, the generated frames should update the camera view, the visible body, nearby people, and the persistent scene layout in a mutually consistent way. These properties quickly expose the limits of control based only on text prompts, reference images, or other 2D visual signals.

Recent work addresses important parts of this problem. Explicit camera-control methods improve viewpoint steering \cite{he2024cameractrl, wang2024motionctrl, watson2024controlling,yang2024directavideo, hou2025camtrol}; 3D-informed generators use geometry or spatial memory to preserve world consistency under viewpoint changes~\cite{ren2025gen3c, ren2025cosmosdrivedream, wu2025spmem, huang2025voyager, cao2025uni3c, yu2025contextasmemory, li2025vmem}; and egocentric models show that body pose is a powerful signal for predicting first-person visual futures~\cite{bai2025peva,pallotta2025egocontrol,xiu2025egotwin}.
Yet the key ingredients remain disconnected. Existing egocentric pose-conditioned models do not explicitly maintain a persistent memory of the 3D environment, while 3D-consistent video generators generally target generic scenes and do not model ego self-occlusion together with nearby people. As a result, fully controllable egocentric generation that is both geometrically grounded and human-aware remains unresolved.

We introduce \coolname{}, a controllable egocentric video generation model that unifies 3D \underline{E}nvironmental memory with \underline{E}go--\underline{E}xo human pose \underline{C}ontrol, as illustrated in \cref{fig:overview}.
Our key idea is to build structured and compact conditions that disentangle persistent scene structure from human-driven dynamics. We represent the static world with a semi-dense point cloud-based 3D memory reconstructed from context frames, and render this memory into the target future viewpoints to obtain conditioning aligned with the target future frames.
The semi-dense point clouds can be obtained from simultaneous localization and mapping (SLAM) or structure-from-motion (SfM) algorithms, which reconstruct 3D points in high-gradient image regions.
To further allow the model to infer accurate image texture, we augment each 3D point with appearance descriptors from video-VAE features.
In indoor egocentric scenes, dynamics are largely human-driven, so we explicitly control ego--exo motion and leave induced object changes to the video model.
To model human dynamics, including both \textit{exo human} dynamics, i.e., other people observed by the wearer, and \textit{ego human} dynamics, i.e., the wearer's own body motion, we render every dynamic 3D skeleton into the conditioning camera viewpoints. Since the wearer's own body is only partially observed in these views, we also introduce ego pose encoding tokens to capture the full (amodal) egocentric body motion.
During generation, 3D memory anchors sparse view-aligned structure, pose controls steer human motion, and the diffusion prior fills missing appearance, people, and dynamics.

We build \coolname{} on a pretrained video diffusion model, VACE~\cite{jiang2025vace, wan2025wan}, fine-tuned and evaluated on Nymeria~\cite{ma2024nymeria}. \coolname{} strongly outperforms 3D-aware and egocentric baselines on visual fidelity, camera-motion adherence, object consistency, and ego/exo human control. Because \coolname{}'s conditioning is explicit, it also supports intuitive scene editing, such as removing objects from the 3D memory or changing the motion of surrounding people.
Our contributions are:
\begin{itemize}[topsep=0pt,itemsep=0pt,parsep=0pt,partopsep=0pt]
    \item We propose \coolname{}, a controllable egocentric video diffusion framework that jointly models the 3D environment and ego and exo human motion.
    \item We introduce a semi-dense point cloud-based 3D environmental memory, augmented with per-point appearance features, enabling geometrically grounded generation under strong egocentric camera motion.
    \item We design an ego pose encoder that converts 3D body joints and wrist poses into persistent cross-attention tokens, improving ego-motion adherence under self-occlusion and out-of-view body configurations.
    \item We demonstrate state-of-the-art results on the Nymeria dataset, and show that explicit memory and pose conditioning enable intuitive scene editing.
\end{itemize}

\vspace{-0.5em}
\section{Related Work}
\label{sec:related}

\noindent\textbf{Video Generation with Camera and Pose Controls.}
A broad family of controllable video generators improves adherence to user intent by conditioning diffusion models on explicit motion cues, including camera trajectories and articulated poses.
Recent work has made camera motion a first-class control signal for text/image-to-video generation, using learned controllers or plug-and-play guidance to follow user-specified paths \cite{he2024cameractrl,wang2024motionctrl,yang2024directavideo,hou2025camtrol}.
To reduce drift under viewpoint changes, several methods further inject geometry-aware camera conditioning---especially in image-to-video and diffusion-transformer settings---leading to stronger 3D-consistent camera control \cite{xu2024camco,zheng2024cami2v,li2025realcami2v,bahmani2025vd3d,bahmani2025ac3d}.
Meanwhile, unified motion control frameworks begin to jointly control camera and actor/object motion using 3D or 6D pose representations \cite{cao2025uni3c,shuai2025freeform}.
However, most existing controllers target exocentric or generic domains and do not explicitly handle egocentric self-occlusion and the tight coupling between the wearer's body motion and camera motion. \coolname{} addresses this gap by conditioning on exo 2D skeleton renderings and ego 3D body and wrist poses, and introducing pose tokenization for persistent cross-attention that preserves ego human control even when the body is partially out of view.

\noindent\textbf{3D-aware Video Generation and Spatial Memory.}
Maintaining long-horizon spatial consistency under large viewpoint changes remains challenging for video diffusion models, and recent methods increasingly inject explicit 3D structure into the generation process.
One line of work conditions video generators on camera and geometric cues---through camera-controllable architectures or explicit 3D supervision \cite{zhang2025wvd}---or through rendering-based conditioning from reconstructed scene representations \cite{ren2025gen3c,huang2025voyager,wang2025unposedphotos}.
In particular, Gen3C \cite{ren2025gen3c} constructs a depth-derived point-cloud cache and renders it along a target trajectory to guide synthesis, improving camera control and mitigating geometric drift.
Complementary to one-shot caches, spatial memory mechanisms maintain an explicit world state over time, retrieving and updating geometry to avoid attending to the full history \cite{li2025vmem,wu2025spmem,zhao2025spatia,wang2025evoworld,lee2025scene}.
Related efforts promote 3D consistency via multi-view video generation for 4D content \cite{voleti2024sv3d,xie2024sv4d,zhang2024fourdiffusion} or training-time alignment of diffusion representations with geometric priors \cite{danier2025vicodr,du2026videogpa}.
While these advances focus on scene consistency and camera control, they typically assume a free-flying camera and do not explicitly target egocentric body control or multi-human interactions.
\coolname{} adapts the 3D-memory paradigm to egocentric generation by building a semi-dense point-based map enriched with per-point appearance features from video latents, providing spatiotemporally aligned, viewpoint-aware conditioning for future synthesis.

\noindent\textbf{Egocentric Video Generation and Ego--Exo Settings.}
Recent progress in egocentric video generation has been driven by large-scale data and diffusion-based generators, with benchmarks such as EgoVid-5M providing paired video--action supervision and strong baselines for controllable ego synthesis~\cite{wang2025egovid5m}.
Building on these resources, egocentric ``world models'' increasingly condition future prediction on actions or kinematics: PEVA uses whole-body pose trajectories for egocentric video prediction~\cite{bai2025peva}, EgoControl enables pose-controllable egocentric generation via 3D full-body signals~\cite{pallotta2025egocontrol}, and EgoTwin explores joint generation of egocentric video and motion~\cite{xiu2025egotwin}; complementary work studies long-horizon consistency via long-context diffusion mechanisms~\cite{zhang2025egolcd}.
In parallel, ego--exo settings investigate cross-view generation and translation, including exo-to-ego synthesis and extensions~\cite{luo2024putmyself,liu2024exo2ego,kang2025egox,mahdi2025exo2egosyn}, as well as ego-to-exo generation~\cite{luo2024intention} and cross-view ego future prediction from exo observations~\cite{xu2025egoexo_gen}.
Different from translation-centric formulations or pose-only conditioning, \coolname{} targets future egocentric generation from past observations while \emph{jointly} leveraging explicit 3D memory and controllable ego--exo pose signals, aiming to improve both scene coherence and motion adherence in complex real-world egocentric recordings.

\vspace{-0.5em}
\section{Method}
\label{sec:method}

\newcommand{\spamem}{\mathcal{M}}
\newcommand{\poseego}{\mathbf{p}^e}
\newcommand{\poseegoskel}{\mathbf{p}^{e,s}}
\newcommand{\poseegowrist}{\mathbf{p}^{e,w}}
\newcommand{\poseexo}{\mathbf{p}^x}
\newcommand{\pcd}{\mathbf{P}}
\newcommand{\pcdcolor}{\mathbf{C}}
\newcommand{\pcdfeat}{\mathbf{F}}
\newcommand{\featmap}{\mathbf{e}}
\newcommand{\imgh}{H}
\newcommand{\imgw}{W}
\newcommand{\latenth}{\imgh'}
\newcommand{\latentw}{\imgw'}
\newcommand{\latentdim}{D}
\newcommand{\cond}{\mathcal{C}} %
\newcommand{\condpc}{\mathbf{c}^{\text{pc}}} %
\newcommand{\condfeat}{\mathbf{c}^{\text{feat}}} %
\newcommand{\condvid}{\mathbf{c}^{\text{ctrl}}} %

\begin{figure}[t]
    \centering
    \includegraphics[width=\textwidth,trim={0in 7.9in 4.8in 0},clip]{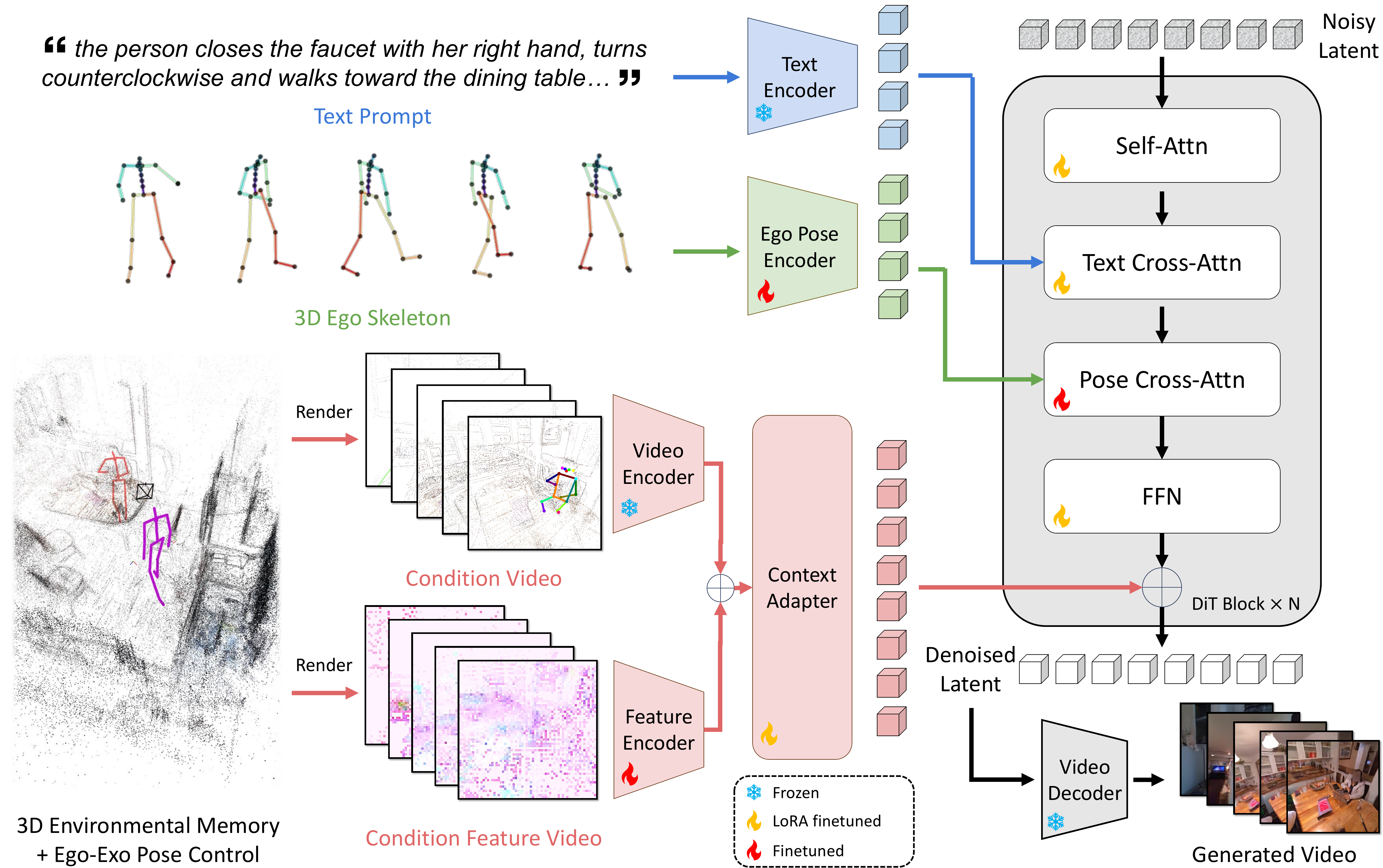}
    \vspace{-1.5em}
    \caption{Overview of \coolname{}. A latent video DiT iteratively denoises noisy video latents conditioned on a text prompt, view-aligned renderings of a semi-dense point cloud-based 3D memory augmented with VAE appearance features, and ego--exo human pose controls. The rendered memory is injected via the context adapter, while pose controls are provided both as pose drawings in the conditioning video and as persistent ego pose tokens for cross-attention.\vspace{-1.5em}}
    \label{fig:method-pipeline}
\end{figure}

\coolname{} conditions a latent video diffusion transformer (DiT) \cite{peebles2023dit} on view-aligned renderings of a 3D environmental memory and on ego--exo human pose controls, as illustrated in \cref{fig:method-pipeline}.
In this section, we first introduce the problem setup (\cref{sec:setup}), and then describe the latent diffusion backbone (\cref{sec:diffusion}), the 3D memory and view-aligned rendering (\cref{sec:method-memory}), and finally ego body pose conditioning (\cref{sec:method-pose}).

\vspace{-0.5em}
\subsection{Problem Overview}
\label{sec:setup}

We study conditional egocentric video generation. Given $m$ context RGB frames $\mathbf{x}_{1:m} \in \mathbb{R}^{m \times \imgh \times \imgw \times 3}$ and a text prompt $\tau$, we synthesize $n$ frames $\widehat{\mathbf{y}}_{1:n} \in \mathbb{R}^{n \times \imgh \times \imgw \times 3}$ that are consistent with the observed environment and follow user-specified human motion controls.
The corresponding GT target frames are denoted as $\mathbf{y}_{1:n} \in \mathbb{R}^{n \times \imgh \times \imgw \times 3}$.
The context frames are often past observations, but can be any set of relevant frames.
From the context clip, we construct a 3D environmental memory $\spamem$ and render it along a target camera trajectory $\mathbf{T}_{1:n}$ to condition our generative model on the 3D environment and the camera's path through it.
In egocentric videos, $\mathbf{T}_{1:n}$ is typically given by the future headset pose trajectory; at inference, it can be user-specified.
We control human motion with ego motion $\poseego_{1:n}$ and exo motion $\poseexo_{1:n}$: ego human motion consists of 3D body joints $\poseegoskel_{1:n}$ and 6DoF wrist poses $\poseegowrist_{1:n}$, and exo human motion is specified as 2D skeleton maps in the image plane.
We additionally condition our model on the last observed frame $\mathbf{x}_m$, serving as a high-detail visual cue. %
Overall, we learn:
\begin{equation}
p_\theta(\widehat{\mathbf{y}}_{1:n}\mid \mathbf{x}_{m}, \tau, \spamem, \mathbf{T}_{1:n}, \poseego_{1:n}, \poseexo_{1:n}).
\end{equation}

\vspace{-1em}
\subsection{Latent Video Diffusion Preliminaries}
\label{sec:diffusion}

\coolname{} builds on VACE~\cite{wan2025wan,jiang2025vace}, a pretrained latent video diffusion model.
Let $\mathcal{E}(\cdot)$ and $\mathcal{D}(\cdot)$ denote a pretrained video VAE encoder and decoder.
We encode the target clip into latents $\mathbf{z}_0=\mathcal{E}(\mathbf{y}_{1:n})$ and apply a flow-matching noising process~\cite{lipman2022flow,esser2024rectifiedflow}:
\begin{equation}
\mathbf{z}_t = (1-\sigma_t)\,\mathbf{z}_0 + \sigma_t\,\boldsymbol{\epsilon},
\quad \boldsymbol{\epsilon}\sim\mathcal{N}(\mathbf{0},\mathbf{I}), \; t\in\{1,\dots,T\}.
\label{eq:forward}
\end{equation}
Here $\sigma_t\in[0,1]$ is a monotonically increasing noise schedule over timesteps, and the corresponding flow-matching target is $\mathbf{u} = \boldsymbol{\epsilon}-\mathbf{z}_0$.
A denoiser $\epsilon_\theta$ predicts the flow-matching target conditioned on text and control signals:
\begin{equation}
\widehat{\mathbf{u}} = \epsilon_\theta(\mathbf{z}_t, t;\, \tau, \cond),
\label{eq:denoise}
\end{equation}
where $\epsilon_\theta$ is implemented as a video DiT~\cite{peebles2023dit}.
Here $\cond$ aggregates all non-text conditions, including the last context frame $\mathbf{x}_m$, view-aligned 3D memory renderings (\cref{sec:method-memory}), and ego--exo pose controls (\cref{sec:method-pose}).
We train with a timestep-weighted regression objective:
\begin{equation}
\mathcal{L} = \mathbb{E}_{\mathbf{z}_0,t,\boldsymbol{\epsilon}}
\left[w(t)\,\left\| (\boldsymbol{\epsilon}-\mathbf{z}_0) - \epsilon_\theta(\mathbf{z}_t,t;\tau,\cond)\right\|_2^2\right].
\label{eq:loss}
\end{equation}
At inference, we sample from the reverse process starting from $\mathbf{z}_T\sim\mathcal{N}(\mathbf{0},\mathbf{I})$
and decode the final latent clip via $\mathcal{D}(\cdot)$.
As shown in \cref{fig:method-pipeline}, text conditioning and ego pose conditioning enter through cross-attention,
and view-aligned spatiotemporal conditions are injected through a context adapter network~\cite{jiang2025vace}.

\vspace{-0.5em}
\subsection{3D Spatial Memory for Consistent Video Generation}
\label{sec:method-memory}

Rapid egocentric camera motion makes it difficult for video diffusion models to preserve a coherent 3D world over time.
We address this by constructing a 3D spatial memory $\spamem$ from the context and rendering it into the target viewpoints as view-aligned conditioning.
Concretely, we represent $\spamem$ as a semi-dense point cloud $\pcd$ with per-point color $\pcdcolor$ and latent appearance descriptors $\pcdfeat$.

\noindent{\bf Memory construction.}
Given the $m$ context frames $\mathbf{x}_{1:m}$, we run a SLAM algorithm~\cite{engel2023aria} to build a semi-dense point cloud $\pcd \in \mathbb{R}^{N\times 3}$ of the environment.
For each point in $\pcd$, we assign an RGB color $\pcdcolor \in \mathbb{R}^{N\times 3}$ by sampling from the context frames where the point is observed.
The point cloud is voxel-subsampled, and the color of each voxel is computed by mean pooling the colors of points within the voxel.
Compared to video models that rely on monocular dense depth estimation~\cite{ren2025gen3c, cao2025uni3c, wu2025spmem}, our semi-dense reconstruction provides metrically accurate point clouds and a temporally coherent semi-dense map.
However, semi-dense SLAM reconstructs points primarily at high image-gradient locations (e.g., edges and textured regions), so the resulting point cloud concentrates on visually informative structure while leaving low-texture areas under-represented. See \cref{fig:method-pipeline} for an example 3D memory.

\noindent{\bf Per-point appearance representation.}
The semi-dense map $\spamem$ is sparse and may miss appearance details in textureless surfaces or partially occluded areas.
To enrich this spatial memory, we augment each 3D point with local appearance features extracted from the input frames.
We encode the context frames $\mathbf{x}_{1:m}$ using the video VAE encoder $\mathcal{E}$ (Sec.~\ref{sec:diffusion}), and extract from these feature maps per-point latent descriptors $\pcdfeat \in \mathbb{R}^{N\times \latentdim}$. %
Similar to the construction process of $\pcd$ and $\pcdcolor$, we project each 3D point into the context frames' feature maps, sample latent descriptors using bilinear interpolation, and aggregate multi-view samples by voxelizing the points and mean-pooling features within each voxel.
During generation, these learned appearance features allow the diffusion model to extrapolate texture and fill regions not directly covered by the sparse geometry, while remaining grounded by the high-gradient structural support of the SLAM map.

\noindent{\bf View-aligned rendering and conditioning.}
Given a target camera trajectory $\mathbf{T}_{1:n}$, we project and render the augmented point cloud (with both color $\pcdcolor$ and latent features $\pcdfeat$) into each target viewpoint.
This produces a sequence of RGB renderings $\condpc$ and rendered feature maps $\condfeat$ that are spatiotemporally aligned with the target video frames $\mathbf{y}_{1:n}$. Human pose conditions are also drawn on $\condpc$, resulting in $\condvid$, as will be detailed in \cref{sec:method-pose}.
We then encode $\condvid$ using the video encoder, yielding latent features $\featmap^{\text{ctrl}} \in \mathbb{R}^{n' \times \latenth \times \latentw \times \latentdim}$.
In parallel, we encode $\condfeat$ using a separate feature encoder to obtain $\featmap^{\text{feat}} \in \mathbb{R}^{n' \times \latenth \times \latentw \times \latentdim}$.
We fuse the two streams by summation, $\featmap = \featmap^{\text{ctrl}} + \featmap^{\text{feat}}$, and feed the fused features $\featmap$ through the VACE Context Adapter~\cite{jiang2025vace} to inject them into the video DiT blocks as residual conditioning (\cref{fig:method-pipeline}).
Note that \coolname{} does not take visibility or depth maps associated with $\featmap$ as explicit inputs, but learns to reason about them implicitly.

\vspace{-0.5em}
\subsection{Controllable Video Generation with Ego-Exo Human Poses}
\label{sec:method-pose}

Human motion drives egocentric video dynamics, inducing camera motion, self-occlusions, and interactions with the environment.
To enable fine-grained control over human motion in the generated videos, \coolname{} conditions on pose sequences for both the ego human $\poseego$ and the exo humans $\poseexo$ over the target $n$ frames and is trained to follow these controls.

\begin{figure}[t]
    \centering
    \includegraphics[width=\textwidth,trim={0in 10.3in 3.4in 0},clip]{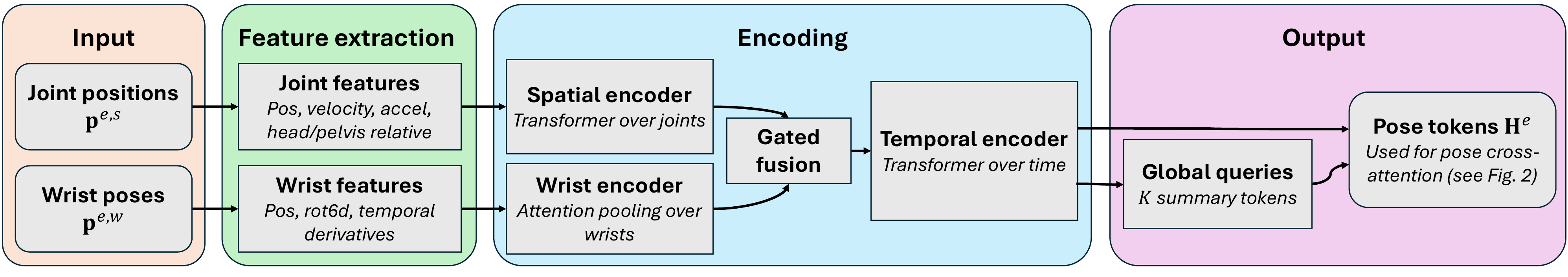}
    \vspace{-2em}
    \caption{Ego pose encoder. The encoder maps ego 3D body joints $\poseegoskel$ and 6DoF wrist poses $\poseegowrist$ to motion-aware tokens $\mathbf{H}^e$ using spatial and temporal transformers and gated fusion. $\mathbf{H}^e$ is then used for pose cross-attention (\cref{fig:method-pipeline}), providing persistent ego human motion control when the body is out of view or occluded. See the appendix for network details.\vspace{-2em}}
    \label{fig:ego-encoding-scheme}
\end{figure}

\noindent\textbf{Exo human pose control.}
Exo humans are typically fully visible and appear at a moderate distance from the camera wearer, so coarse articulated motion provides a strong control signal.
Following prior work~\cite{zhang2023controlnet, cao2025uni3c, jiang2025vace}, we rasterize their 2D skeletons and overlay them on the point-cloud renderings $\condpc$ to form the RGB conditioning video $\condvid$. See example conditioning frames in \cref{fig:method-pipeline}.

\noindent\textbf{Ego human pose control.}
The ego wearer is often only partially observed due to being out of view or close-range self-occlusion, yet their motion is tightly coupled to the camera trajectory and dominates first-person visual dynamics. We condition on target ego 3D body joints $\poseegoskel \in \mathbb{R}^{n\times J \times 3}$ and 6DoF wrist poses $\poseegowrist \in \mathbb{R}^{n\times 2 \times 4 \times 4}$, represented as 3D joint positions and $\mathrm{SE}(3)$ transforms, respectively, in the camera coordinate frame. Similar to exo human pose control, we project these 3D signals into each frame using the camera intrinsics and rasterize them into $\condvid$ as 2D skeletons and small coordinate axes. However, because the ego body is frequently out of view or occluded, these drawings can be sparse and provide limited control on their own.

\noindent\textbf{Persistent ego pose tokens.}
To provide ego human pose control even under partial observability, we further encode the full target ego pose sequence into persistent tokens, as illustrated in~\cref{fig:ego-encoding-scheme}.
Specifically, we augment joint and wrist poses with first-order temporal differences, represent joints by their 3D coordinates and wrists by their positions and a continuous 6D rotation representation \cite{zhou2019continuity}, and linearly project all features into a shared $D$-dimensional space.
We then embed the joint and wrist sequences, aggregate across joints and wrists with spatial transformer blocks \cite{vaswani2017attention} and attention pooling \cite{lee2019settransformer}, fuse the two streams with a gated module \cite{srivastava2015highway, arevalo2017gated}, and propagate information temporally.
This results in $n$ pose tokens, each summarizing the joint and wrist pose information of one target frame.
We additionally introduce $K$ learned summary tokens that cross-attend to the $n$ pose tokens~\cite{carion2020detr}, and concatenate them with the pose tokens.
The resulting $n+K$ tokens $\mathbf{H}^e \in \mathbb{R}^{(n+K)\times D_{DiT}}$ condition the video DiT via a dedicated pose cross-attention, as shown in~\cref{fig:method-pipeline}, where $D_{DiT}$ denotes the video DiT token dimension. More network details are provided in the appendix.

\vspace{-0.5em}
\section{Experiments}
\label{sec:experiment}

We evaluate \coolname{} on the Nymeria egocentric benchmark~\cite{ma2024nymeria} and analyze both generation quality and controllability. We compare against 3D-aware methods with explicit spatial representations, conditional video diffusion models that take aligned control videos as input, and egocentric pose-conditioned generators. Beyond standard video-quality metrics, we report camera-motion accuracy, object consistency, and human pose adherence.
We analyze each conditioning component via ablations and show video generation results with controllable edits.

\vspace{-0.5em}
\subsection{Experiment Setup}

We first detail the experimental setup used to evaluate \coolname{} and baselines.

\noindent\textbf{Dataset.}
We use the Nymeria dataset~\cite{ma2024nymeria}, which contains over 1{,}100 egocentric recordings captured with Project Aria glasses~\cite{engel2023aria} across diverse indoor and outdoor environments. Each recording includes dense natural-language captions, time-synchronized 3D body motion from an Xsens inertial motion capture system~\cite{movella_xsens_mvn_link}, and 6DoF wrist poses obtained from the miniAria wristbands. We use Project Aria MPS~\cite{aria_mps} to recover camera trajectories and semi-dense point cloud reconstructions that serve as the 3D memory input.
Each recording session includes a \textit{participant} who performs the activity, an \textit{observer}, and additional bystanders. We treat the participant as the ego wearer and use the participant recordings for training and evaluation. Since bystanders lack synchronized 3D pose annotations, we detect exo humans with YOLO-X~\cite{ge2021yolox} and extract 2D skeletons using ViTPose~\cite{xu2022vitpose}; these skeletons are used as exo human pose controls. We adopt the train/validation/test split from~\cite{guzov2025hmd2} with disjoint environments and participant identities across splits.

\noindent\textbf{Baselines.}
We compare \coolname{} against baselines that cover complementary points in the design space. Splatfacto~\cite{splatfacto2023} is a Gaussian Splatting novel-view synthesis method that reconstructs a scene from context frames and renders target viewpoints. We also evaluate two 3D-aware video generators that use explicit spatial memory, VMem~\cite{li2025vmem} and Gen3C~\cite{ren2025gen3c}. As a strong conditional video diffusion baseline, we include VACE~\cite{jiang2025vace}, which takes spatiotemporally aligned conditioning videos as input. We report VACE results for the released pretrained checkpoint and for models fine-tuned on Nymeria under three conditioning settings: exo human pose only, point-cloud renderings only, and point-cloud renderings with exo human pose. Finally, we compare to egocentric video generators trained and evaluated on Nymeria, including PEVA~\cite{bai2025peva}, EgoControl~\cite{pallotta2025egocontrol}, and EgoTwin~\cite{xiu2025egotwin}.

\noindent\textbf{Implementation Details.}
We preprocess Nymeria videos by devignetting and rectifying them to a pinhole camera model~\cite{gu2024egolifter}, then subsample to 10 FPS and resize frames to $512\times 512$. We segment each recording into 280-frame snippets with a stride of 10 seconds. The first $m=200$ frames form the context $\mathbf{x}_{1:m}$ for constructing the 3D memory, and the remaining $n=80$ frames define the target clip $\mathbf{y}_{1:n}$. Notably, the final context frame $\mathbf{x}_m$ is also provided as input to \coolname{}, serving as a visual cue for prediction.
We initialize \coolname{} from \texttt{Wan2.1-VACE-1.3B}~\cite{jiang2025vace, wan2025wan}, but the proposed framework is compatible with any video latent diffusion model. We LoRA fine-tune the main video DiT and the context adapter with rank 256~\cite{hu2022lora}. The ego pose encoder, feature encoder, and pose cross-attention layers are trained from scratch, as shown in \cref{fig:method-pipeline}. \coolname{} is trained using the Adam optimizer~\cite{kingma2014adam} with a learning rate of $1\times 10^{-4}$ for 50 epochs on 64 NVIDIA H100 GPUs, with a per-device batch size of 1.
At inference, we use a fixed 50-step FlowMatch sampling schedule~\cite{lipman2022flow} with shift parameter $\sigma_{\text{shift}} = 5.0$ and classifier-free guidance scale $\omega = 5.0$.

\vspace{-0.5em}
\subsection{Evaluation Metrics}

{\bf Visual Quality and Fidelity.}
We report standard video generation metrics: Fr\'{e}chet Video Distance (FVD)~\cite{unterthiner2018fvd} for overall visual and temporal fidelity.
Since generation is conditioned on the 3D spatial memory and camera poses, the predicted frames are expected to be pixel-aligned with the ground-truth frames. Therefore, we also evaluate LPIPS~\cite{zhang2018lpips} for perceptual similarity, and PSNR/SSIM~\cite{wang2004ssim} for pixel-level reconstruction and structural consistency.

\noindent{\bf Object Consistency.}
To measure semantic object consistency, we run the open-vocabulary detector OWLv2~\cite{minderer2023owlv2} on both ground-truth and generated videos. We match predicted boxes to ground-truth boxes with the Hungarian algorithm~\cite{kuhn1955hungarian} and report Obj-F1 and Obj-mIoU.

\noindent{\bf Camera Motion Control.}
We evaluate camera-motion adherence by treating Project Aria MPS poses~\cite{aria_mps} as ground-truth trajectories. We estimate trajectories from generated videos with VIPE~\cite{huang2025vipe}, align them to ground truth using a Sim(3) transform, and report translation error (TErr) and rotation error (RErr) following prior work~\cite{he2024cameractrl,pallotta2025egocontrol}.

\begin{table}[t]
  \newcommand{\raf}[1]{\textbf{#1}}
  \newcommand{\ras}[1]{\underline{#1}}
  \centering
  \caption{Quantitative evaluation on the Nymeria dataset. We indicate whether each model is fine-tuned on the Nymeria training set (``FT''). The \raf{best} and \ras{second-best} results for each metric are bold and underlined, respectively.\vspace{-1em}}
  \label{tab:quant-nymeria}
  \resizebox{\textwidth}{!}{%
    \begin{tabular}{lc|cccc|cc|cc|cc|cc}
      \toprule
      \multicolumn{2}{r|}{Metric} & \multicolumn{4}{c|}{Visual Quality} & \multicolumn{2}{c|}{Camera Control} & \multicolumn{2}{c|}{Object} & \multicolumn{2}{c|}{Exo Human Control} & \multicolumn{2}{c}{Ego Human Control} \\
      \cmidrule(lr){3-6} \cmidrule(lr){7-8} \cmidrule(lr){9-10} \cmidrule(lr){11-12} \cmidrule(lr){13-14}
      \multicolumn{1}{l}{Method} & FT & FVD$\downarrow$ & LPIPS$\downarrow$ & PSNR$\uparrow$ & SSIM$\uparrow$ & TErr(cm)$\downarrow$ & RErr($\degree$)$\downarrow$ & Obj-F1$\uparrow$ & Obj-mIoU$\uparrow$ & Exo-F1$\uparrow$ & PCK@10\%$\uparrow$ & Hand-F1$\uparrow$ & Hand-mIoU$\uparrow$  \\
      \midrule
      Splatfacto~\cite{splatfacto2023} & $-$ & 1040 & 0.494 & \ras{17.8} & \ras{0.620} & 6.60 & \raf{21.5} & 41.05 & 57.42 & 15.28 & 63.13 & 21.28 & \ras{19.28} \\
      Gen3C~\cite{ren2025gen3c} & $\times$ & 974 & 0.565 & 15.8 & 0.541 & 21.37 & 88.5 & 25.66 & 42.00 & 12.39 & 42.08 & 25.54 & 14.50 \\
      VMem~\cite{li2025vmem} & $\times$ & 465 & 0.618 & 12.7 & 0.470 & 16.82 & 113.0 & 14.74 & 32.45 & 8.67 & 2.55 & 26.31 & 17.68 \\
      VACE~\cite{jiang2025vace} (exo) & $\times$ & 1180 & 0.644 & 11.7 & 0.401 & 19.00 & 125.0 & 12.43 & 27.37 & 11.14 & 34.69 & 31.92 & 11.28 \\
      VACE~\cite{jiang2025vace} (pts) & $\times$ & 2460 & 0.675 & 11.5 & 0.357 & 14.22 & 72.2 & 6.30 & 18.30 & 3.40 & 85.75 & 17.58 & 15.47 \\
      VACE~\cite{jiang2025vace} (pts+exo) & $\times$ & 2440 & 0.675 & 11.6 & 0.359 & 13.67 & 65.3 & 6.37 & 18.49 & 4.36 & \ras{71.26} & 17.82 & 15.46 \\
      \hline
      VACE~\cite{jiang2025vace} (exo) & $\checkmark$ & 496 & 0.584 & 13.0 & 0.454 & 8.77 & 101.0 & 15.83 & 34.19 & 36.02 & 64.95 & 32.74 & 13.20 \\
      VACE~\cite{jiang2025vace} (pts) & $\checkmark$ & 329 & \ras{0.440} & 17.1 & 0.604 & \ras{3.25} & 24.2 & 46.95 & 71.14 & 19.87 & 37.51 & 34.18 & 18.82 \\
      VACE~\cite{jiang2025vace} (points+exo) & $\checkmark$ & \ras{328} & \ras{0.440} & 17.3 & 0.604 & 5.41 & 24.5 & \ras{47.95} & \ras{71.57} & \ras{36.26} & 64.08 & \ras{34.24} & 18.17 \\
      \hline
      \coolname{} (Ours) & $\checkmark$ & \raf{249} & \raf{0.415} & \raf{18.6} & \raf{0.629} & \raf{2.40} & \ras{23.3} & \raf{52.85} & \raf{74.11} & \raf{39.59} & \raf{72.88} & \raf{41.44} & \raf{24.76} \\
      \bottomrule
      \vspace{-2em}
    \end{tabular}
  }
\end{table}

\noindent{\bf Ego and Exo Human Pose Control.}
We evaluate pose adherence for both the ego wearer and exo people. For ego human control, we follow~\cite{xiu2025egotwin,pallotta2025egocontrol} and report Hand-F1 and Hand-mIoU. We segment ego hands in both ground truth and generated videos using SAM3~\cite{carion2025sam3}. Hand-F1 measures hand presence per frame, and Hand-mIoU measures mask overlap over frames where hands are present in the ground truth. For exo human control, we extract 2D bounding boxes and keypoints from both ground truth and generated videos using YOLO-X and ViTPose~\cite{ge2021yolox,xu2022vitpose}. We match people using Hungarian matching~\cite{kuhn1955hungarian} and report Exo-F1 and PCK@10\%, where PCK (percentage of correct keypoints) uses a threshold of 10\% of the square root of the matched bounding box area. Note that PCK@10\% is computed only over matched bounding boxes, and therefore can remain high even when Exo-F1 is low. As such, Exo-F1 and PCK@10\% should be interpreted jointly for a complete assessment of exo pose control.

\begin{table}[t]
  \centering
  \caption{Comparison with egocentric video-generation baselines on Nymeria. Left: we follow the data split and evaluation protocol used in PEVA~\cite{bai2025peva} and EgoControl~\cite{pallotta2025egocontrol} for a fair comparison. Right: EgoTwin~\cite{xiu2025egotwin} results are reported from the original paper, since the split and implementation details are not available; the protocols may differ.\vspace{-1em}}
  \label{tab:peva_comparison}
  \resizebox{0.4\textwidth}{!}{%
    \begin{tabular}{lccc}
      \toprule
      Model & LPIPS $\downarrow$ & DreamSim $\downarrow$ & FID $\downarrow$ \\
      \midrule
      PEVA XXL~\cite{bai2025peva} & 29.8 & 18.6 & 61.10 \\
      EgoControl~\cite{pallotta2025egocontrol} & 24.3 & 11.3 & 50.68 \\
      \hline
      \coolname{} (Ours) & 17.3 & 10.5 & 50.82 \\
      \bottomrule
    \end{tabular}
  }
  \hspace{1em}
  \resizebox{0.45\textwidth}{!}{%
    \begin{tabular}{lcccccc}
      \toprule
      Model & I-FID$\downarrow$ & FVD$\downarrow$ & TErr(cm)$\downarrow$ & RErr($\degree$)$\downarrow$ \\
      \midrule
      EgoTwin~\cite{xiu2025egotwin} & 98.2 & 1033.5 & 67.0 & 26.4 \\
      \hline
      \coolname{} (Ours) & 31.0 & 249.1 & 2.40 & 23.3 \\
      \bottomrule
    \end{tabular}
  }
  \vspace{-0.5em}
\end{table}

\begin{table}[t]
\centering
\caption{Ablations on conditioning inputs and proposed components. Starting from the \textit{Base} model~\cite{jiang2025vace} conditioned only on the initial image and text, we incrementally add rendered SLAM points (\textit{Point}), rendered per-point VAE appearance features (\textit{Feat}), exo human pose control (\textit{Exo}), ego pose drawn on the conditioning frames (\textit{Drawn Ego}), and our encoded ego pose tokens (\textit{Encoded Ego}). Each variant is trained and evaluated independently.\vspace{-1em}}
\label{tab:quant-ablation}
\resizebox{\textwidth}{!}{%
\begin{tabular}{l|cccc|cc|cc|cc|cc}
\toprule
\multicolumn{1}{r|}{Metric} & \multicolumn{4}{c|}{Visual Quality} & \multicolumn{2}{c|}{Camera Control} & \multicolumn{2}{c|}{Object} & \multicolumn{2}{c|}{Exo Human Control} & \multicolumn{2}{c}{Ego Human Control} \\
\cmidrule(lr){2-5} \cmidrule(lr){6-7} \cmidrule(lr){8-9} \cmidrule(lr){10-11} \cmidrule(lr){12-13}
\multicolumn{1}{l|}{Method}  & FVD$\downarrow$ & LPIPS$\downarrow$ & PSNR$\uparrow$ & SSIM$\uparrow$ & TErr(cm)$\downarrow$ & RErr($\degree$)$\downarrow$ & Obj-F1$\uparrow$ & Obj-mIoU$\uparrow$ & Exo-F1$\uparrow$ & PCK@10\%$\uparrow$ & Hand-F1$\uparrow$ & Hand-mIoU$\uparrow$  \\
\midrule
Base (Image+Text) & \cellcolor[RGB]{208,191,212} 1380 & \cellcolor[RGB]{208,191,212} 0.729 & \cellcolor[RGB]{208,191,212} 10.2 & \cellcolor[RGB]{208,191,212} 0.323 & \cellcolor[RGB]{208,191,212} 14.56 & \cellcolor[RGB]{208,191,212} 124.0 & \cellcolor[RGB]{208,191,212} 4.19 & \cellcolor[RGB]{208,191,212} 18.99 & \cellcolor[RGB]{208,191,212} 0.99 & \cellcolor[RGB]{208,191,212} 1.90 & \cellcolor[RGB]{208,191,212} 27.72 & \cellcolor[RGB]{208,191,212} 9.56 \\
+ Point & \cellcolor[RGB]{199,227,226} 329 & \cellcolor[RGB]{199,227,226} 0.440 & \cellcolor[RGB]{199,227,226} 17.1 & \cellcolor[RGB]{199,227,226} 0.604 & \cellcolor[RGB]{217,242,214} 3.25 & \cellcolor[RGB]{204,237,221} 24.2 & \cellcolor[RGB]{199,227,226} 46.95 & \cellcolor[RGB]{199,227,226} 71.14 & \cellcolor[RGB]{199,225,226} 19.87 & \cellcolor[RGB]{199,227,226} 37.51 & \cellcolor[RGB]{200,225,226} 34.18 & \cellcolor[RGB]{199,232,224} 18.82 \\
\hspace{1ex}+ Feat & \cellcolor[RGB]{200,233,223} 310 & \cellcolor[RGB]{223,244,210} 0.424 & \cellcolor[RGB]{215,241,215} 17.9 & \cellcolor[RGB]{217,242,214} 0.617 & \cellcolor[RGB]{241,247,198} 2.67 & \cellcolor[RGB]{215,241,215} 23.5 & \cellcolor[RGB]{223,244,210} 50.83 & \cellcolor[RGB]{223,244,210} 73.10 & \cellcolor[RGB]{199,227,226} 20.92 & \cellcolor[RGB]{199,233,224} 45.07 & \cellcolor[RGB]{200,225,226} 34.26 & \cellcolor[RGB]{199,228,226} 17.72 \\
\hspace{2ex}+ Exo & \cellcolor[RGB]{236,246,202} 266 & \cellcolor[RGB]{236,246,202} 0.420 & \cellcolor[RGB]{236,246,202} 18.3 & \cellcolor[RGB]{229,245,206} 0.622 & \cellcolor[RGB]{199,227,226} 4.30 & \cellcolor[RGB]{199,227,226} 25.9 & \cellcolor[RGB]{250,248,197} 52.76 & \cellcolor[RGB]{229,245,206} 73.32 & \cellcolor[RGB]{248,248,197} 38.40 & \cellcolor[RGB]{247,248,197} 69.45 & \cellcolor[RGB]{202,220,226} 33.04 & \cellcolor[RGB]{199,227,226} 17.35 \\
\hspace{3ex}+ Drawn Ego & \cellcolor[RGB]{247,248,197} 256 & \cellcolor[RGB]{247,248,197} 0.417 & \cellcolor[RGB]{250,248,197} 18.5 & \cellcolor[RGB]{248,248,197} 0.627 & \cellcolor[RGB]{250,248,197} 2.50 & \cellcolor[RGB]{254,249,200} 21.3 & \cellcolor[RGB]{254,249,200} 53.02 & \cellcolor[RGB]{254,249,200} 74.21 & \cellcolor[RGB]{251,248,198} 38.91 & \cellcolor[RGB]{250,248,197} 70.77 & \cellcolor[RGB]{199,233,224} 36.03 & \cellcolor[RGB]{201,235,223} 19.36 \\
\hspace{4ex}+ Encoded Ego (\coolname{}) & \cellcolor[RGB]{254,249,200} 249 & \cellcolor[RGB]{254,249,200} 0.415 & \cellcolor[RGB]{254,249,200} 18.6 & \cellcolor[RGB]{254,249,200} 0.629 & \cellcolor[RGB]{254,249,200} 2.40 & \cellcolor[RGB]{217,242,214} 23.3 & \cellcolor[RGB]{251,248,198} 52.85 & \cellcolor[RGB]{251,248,198} 74.11 & \cellcolor[RGB]{254,249,200} 39.59 & \cellcolor[RGB]{254,249,200} 72.88 & \cellcolor[RGB]{254,249,200} 41.44 & \cellcolor[RGB]{254,249,200} 24.76 \\
\bottomrule
\end{tabular}
}
\vspace{-1em}
\end{table}

\vspace{-0.5em}
\subsection{Results}
\label{sec:exp-results}

{\bf Quantitative comparison with general-domain video generators.}
We evaluate all methods on 100 snippets from the Nymeria test set and report results in \cref{tab:quant-nymeria}.
Note that the results for object consistency and exo and ego human control metrics are scaled by 100.
For Splatfacto~\cite{splatfacto2023}, we reconstruct a 3D Gaussian Splatting scene for each snippet from the context frames and render the target viewpoints. For Gen3C~\cite{ren2025gen3c} and VMem~\cite{li2025vmem}, we use their official implementations and pretrained weights, feeding the first 200 frames as context. For VACE~\cite{jiang2025vace}, we evaluate both the released pretrained checkpoint and models fine-tuned on Nymeria using three conditioning settings: exo pose only, point cloud only, and point cloud plus exo pose. Each VACE fine-tune run uses the same training setup as \coolname{}.

\newcommand{\qualresultsdata}{%
    000041/{0002,0013,0035,0060,0075},%
    000037/{0019,0029,0042,0069,0081},%
    000023/{0013,0025,0040,0059,0069},%
    000091/{0005,0021,0057,0064,0076},%
}

\begin{figure*}[p]
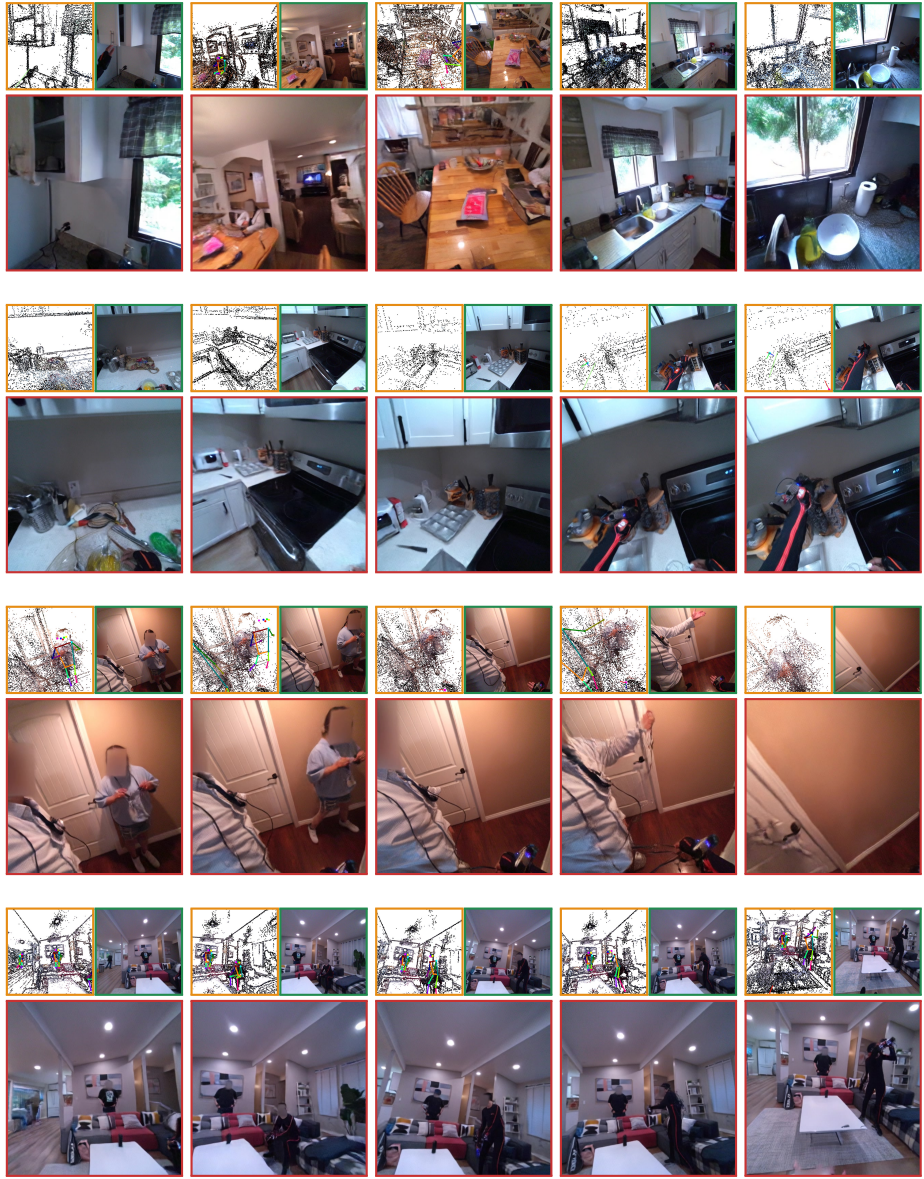

    \centering
    \setlength{\fboxsep}{0pt}
    \setlength{\fboxrule}{0.8pt}
    \foreach \dataidx/\timestamps in \qualresultsdata {%
        \foreach \timestamp in \timestamps {%
            \begin{minipage}[t]{0.192\textwidth}
                \centering
                \fcolorbox{condcolor}{white}{\includegraphics[width=0.475\linewidth]{figures/qual-results/\dataidx/vace/frame_\timestamp.jpg}}%
                \hfill%
                \fcolorbox{gtcolor}{white}{\includegraphics[width=0.475\linewidth]{figures/qual-results/\dataidx/gt/frame_\timestamp.jpg}}\\[0.5pt]
                \fcolorbox{predcolor}{white}{\includegraphics[width=\dimexpr\linewidth-2\fboxrule-2\fboxsep\relax]{figures/qual-results/\dataidx/pred/frame_\timestamp.jpg}}
                \vspace{1pt}
            \end{minipage}\hfill
        }
        \par
    }
    \caption{Qualitative results. Each row corresponds to one video snippet and columns show different timestamps. In each cell, the top row shows the \textcolor{condcolor}{conditioning visualization} (left) and the \textcolor{gtcolor}{ground-truth} frame (right), and the bottom row shows the \textcolor{predcolor}{generated} frame from \coolname{}. \coolname{} maintains consistent 3D scenes (row 1), follows ego-body motion (row 2), and adheres to exo-human motion (rows 3--4). See full videos on the project page.}
    \label{fig:qualresults}
\end{figure*}

Nymeria poses a challenging distribution shift for general-domain video generators due to rapid egocentric motion, frequent self-occlusion, and complex indoor scenes. As shown in \cref{tab:quant-nymeria}, zero-shot models and some 3D-aware baselines degrade substantially. Fine-tuned VACE models recover strong performance when provided with point cloud conditioning, and adding exo skeletons further improves exo-control metrics.
Compared to the best fine-tuned VACE baseline (points+exo), \coolname{} augments points with appearance features, adds the feature encoder branch, and uses the ego pose encoder for persistent ego human control.
These help \coolname{} achieve the strongest overall performance, improving fidelity and object consistency while reducing camera translation error and improving ego hand metrics over the fine-tuned VACE baselines.

Among the non-diffusion baselines, Splatfacto provides accurate viewpoint rendering when the scene is sufficiently reconstructed, but it struggles with dynamic content and regions outside the context views. VMem produces plausible videos but exhibits larger camera-trajectory errors, consistent with the difficulty of transferring to Nymeria's high-frequency egocentric motion.
See qualitative comparisons on the project page.

\noindent{\bf Quantitative comparison with egocentric video generators.}
We also compare against egocentric video generators evaluated on Nymeria in \cref{tab:peva_comparison}. For PEVA~\cite{bai2025peva} and EgoControl~\cite{pallotta2025egocontrol}, we align the data split and follow their evaluation protocol: each method uses a 15-frame context window from the past 3.75 seconds and is evaluated on a single generated frame 2 seconds into the future. We report LPIPS~\cite{zhang2018lpips}, DreamSim~\cite{fu2023dreamsim}, and FID~\cite{fid} in \cref{tab:peva_comparison} (left). The results show that \coolname{} improves LPIPS and DreamSim and achieves comparable FID with the state-of-the-art EgoControl method~\cite{pallotta2025egocontrol}.
For EgoTwin~\cite{xiu2025egotwin}, we cannot fully align the evaluation protocol because the original split and implementation details are not available\footnote{EgoTwin authors informed us they no longer have access to the code.}. We therefore report the numbers from their paper alongside our results as a reference point in \cref{tab:peva_comparison} (right), noting that the protocols may differ.

\begin{figure}[t]
    \centering
    \newlength{\qualablateimgw}%
    \newlength{\qualablatelabelw}%
    \setlength{\qualablatelabelw}{0.3em}%
    \setlength{\qualablateimgw}{\dimexpr(\linewidth-\qualablatelabelw-5.2mm)/6\relax}%
    \newcommand{\vcentered}[1]{\adjustbox{valign=c}{#1}}%

    \begin{tabular}{@{}c@{\hspace{1mm}}c@{\hspace{0.2mm}}c@{\hspace{0.2mm}}c@{\hspace{1mm}}c@{\hspace{0.2mm}}c@{\hspace{0.2mm}}c@{}}
        & \makebox[\qualablateimgw][c]{\scriptsize GT}
        & \makebox[\qualablateimgw][c]{\scriptsize\shortstack{Pred.\ w/}}
        & \makebox[\qualablateimgw][c]{\scriptsize\shortstack{Pred.\ w/o}}
        & \makebox[\qualablateimgw][c]{\scriptsize GT}
        & \makebox[\qualablateimgw][c]{\scriptsize\shortstack{Pred.\ w/}}
        & \makebox[\qualablateimgw][c]{\scriptsize\shortstack{Pred.\ w/o}} \\
        \vcentered{\rotatebox{90}{\scriptsize Feat}} &
        \vcentered{\includegraphics[width=\qualablateimgw]{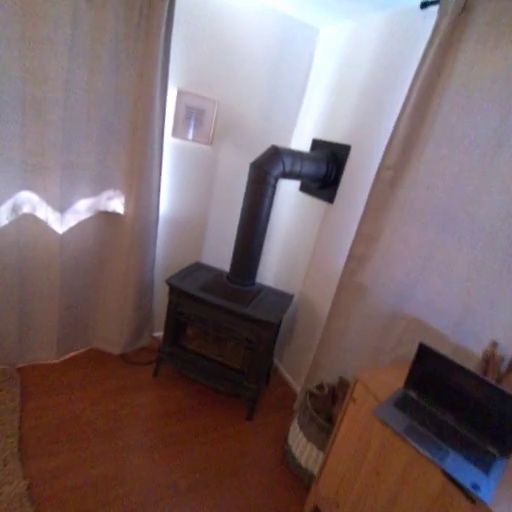}} &
        \vcentered{\includegraphics[width=\qualablateimgw]{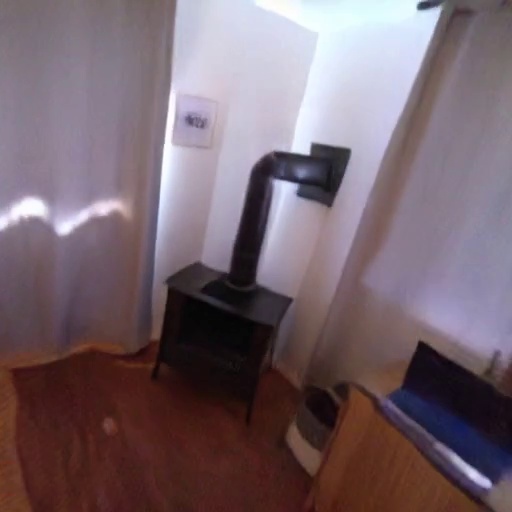}} &
        \vcentered{\includegraphics[width=\qualablateimgw]{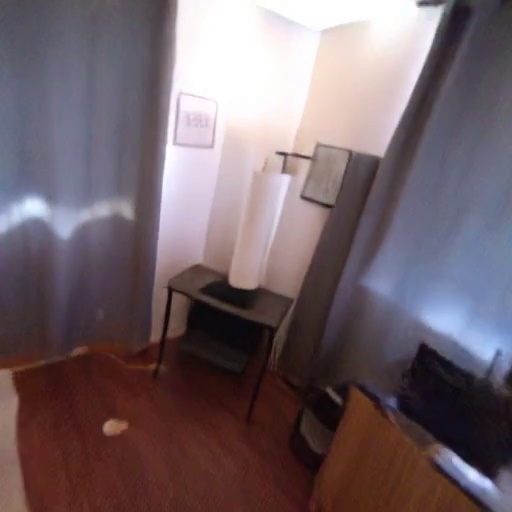}} &
        \vcentered{\includegraphics[width=\qualablateimgw]{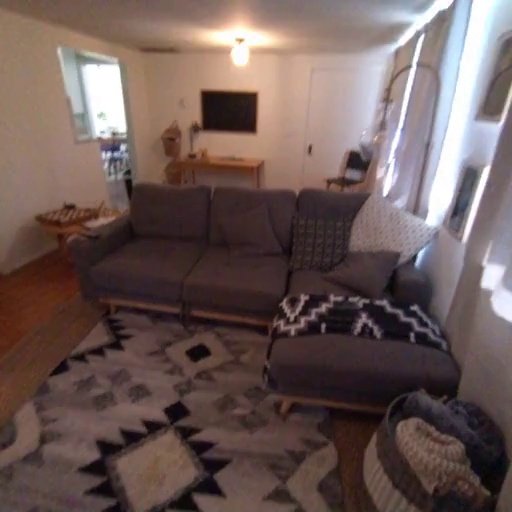}} &
        \vcentered{\includegraphics[width=\qualablateimgw]{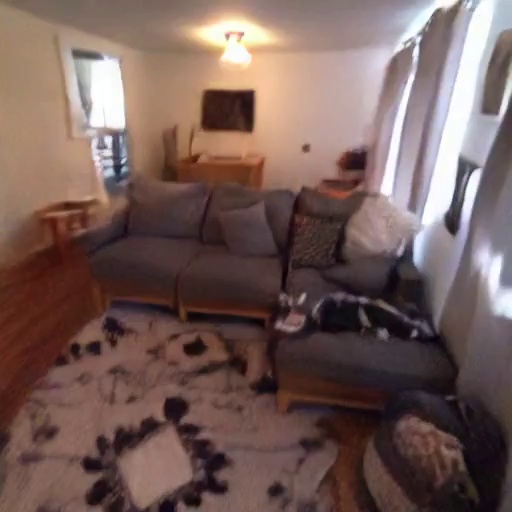}} &
        \vcentered{\includegraphics[width=\qualablateimgw]{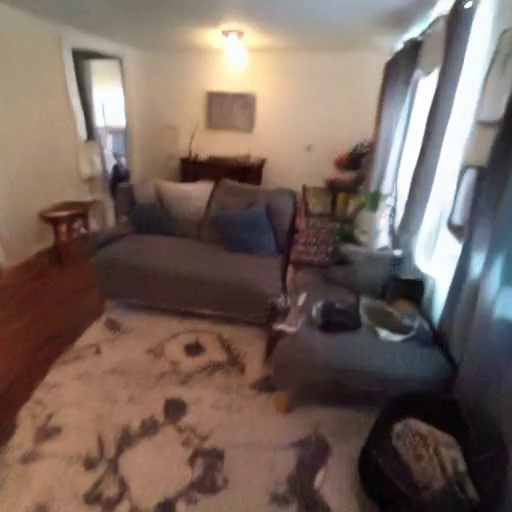}} \vspace{0.5mm}\\

        \vcentered{\rotatebox{90}{\scriptsize Exo}} &
        \vcentered{\includegraphics[width=\qualablateimgw]{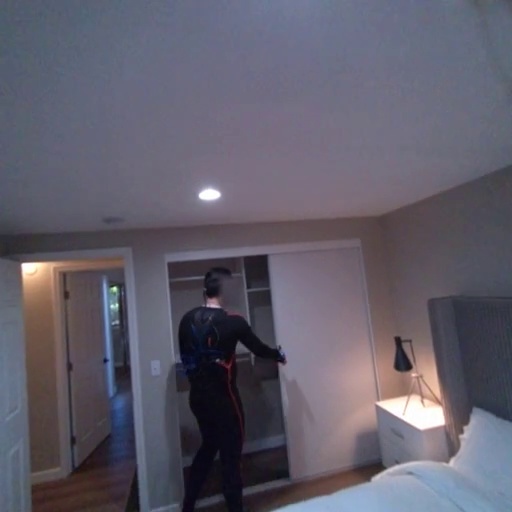}} &
        \vcentered{\includegraphics[width=\qualablateimgw]{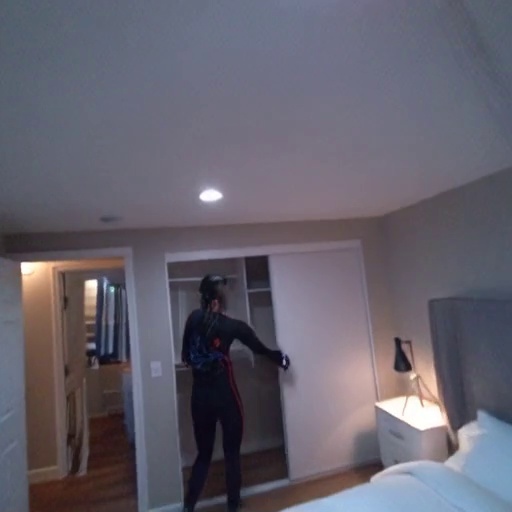}} &
        \vcentered{\includegraphics[width=\qualablateimgw]{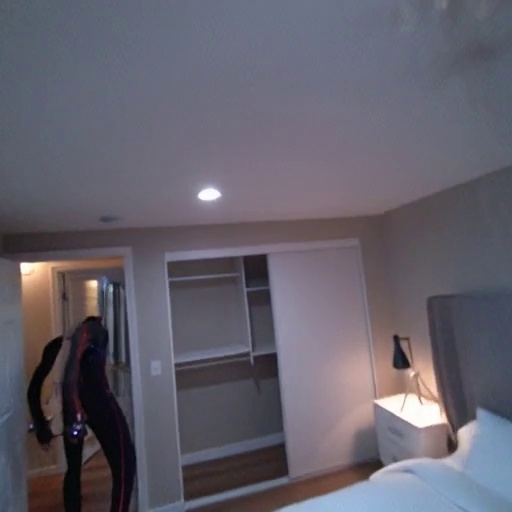}} &
        \vcentered{\includegraphics[width=\qualablateimgw]{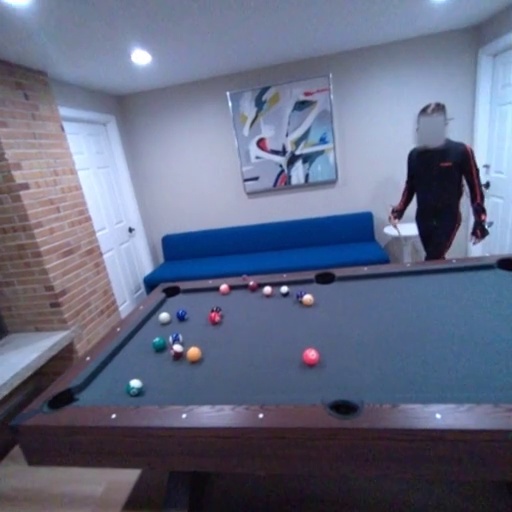}} &
        \vcentered{\includegraphics[width=\qualablateimgw]{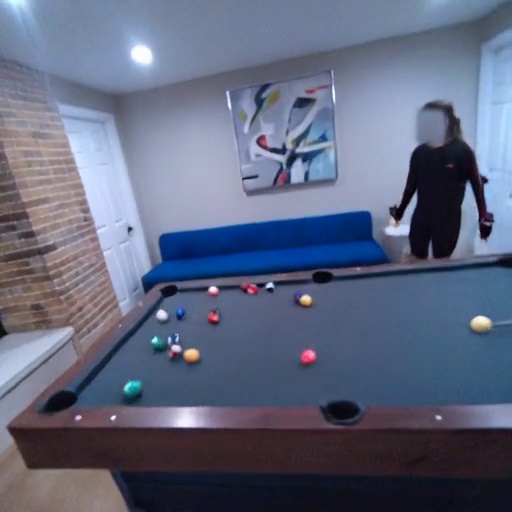}} &
        \vcentered{\includegraphics[width=\qualablateimgw]{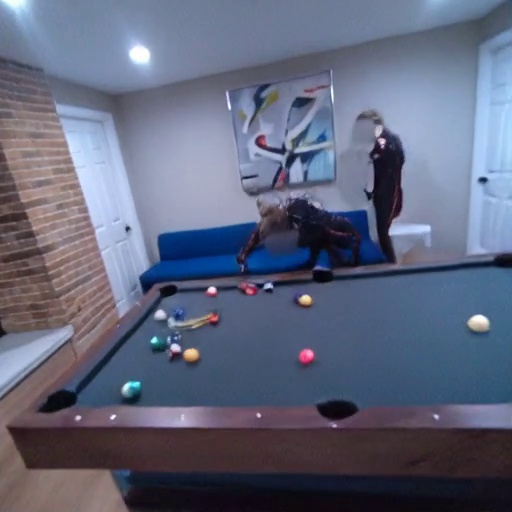}} \vspace{0.5mm}\\

        \vcentered{\rotatebox{90}{\scriptsize Ego}} &
        \vcentered{\includegraphics[width=\qualablateimgw]{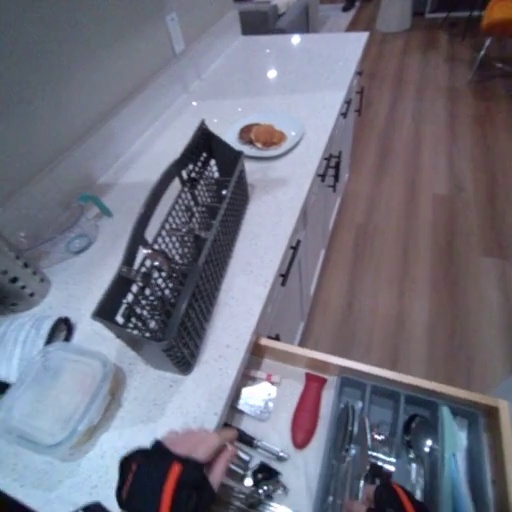}} &
        \vcentered{\includegraphics[width=\qualablateimgw]{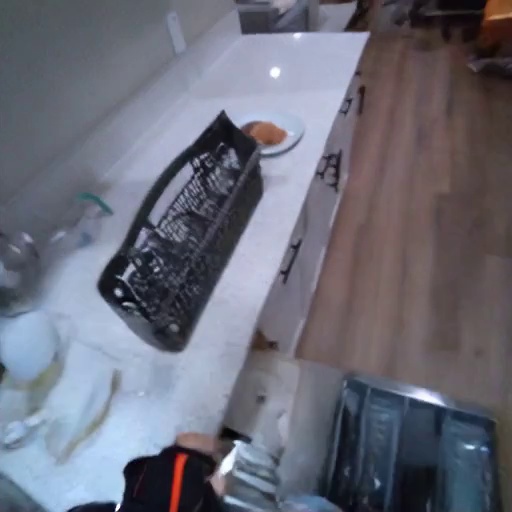}} &
        \vcentered{\includegraphics[width=\qualablateimgw]{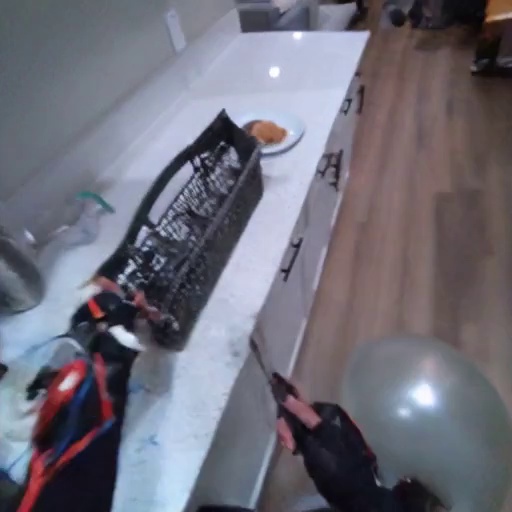}} &
        \vcentered{\includegraphics[width=\qualablateimgw]{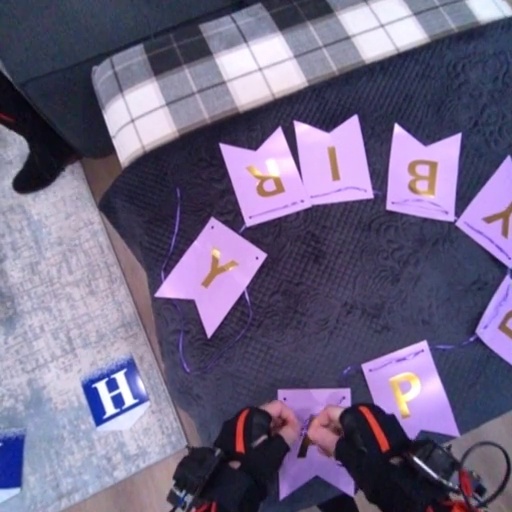}} &
        \vcentered{\includegraphics[width=\qualablateimgw]{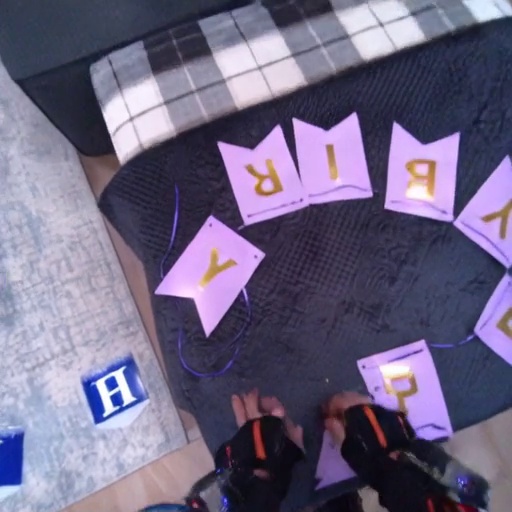}} &
        \vcentered{\includegraphics[width=\qualablateimgw]{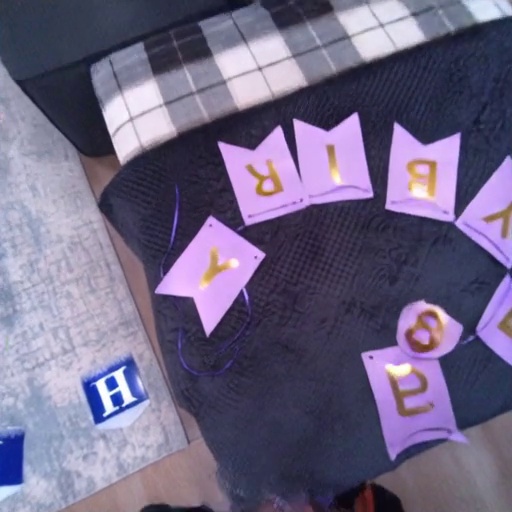} }\\
    \end{tabular}
    \vspace{-0.7em}
    \caption{Qualitative ablations. Each row shows the GT frame and the corresponding predictions with and without each type of condition. Top: removing per-point appearance features (\textit{Feat}) increases texture and color drift. Middle: removing exo pose control (\textit{Exo}) harms exo-motion adherence. Bottom: removing encoded ego pose tokens (\textit{Ego}) reduces ego hand/body fidelity under self-occlusion.\vspace{-2em}}
    \label{fig:qual-ablate}
\end{figure}

\noindent{\bf Qualitative Results.}
\cref{fig:qualresults} visualizes representative generations by \coolname{}. The model maintains coherent 3D scene structure across large viewpoint changes while following both ego actions and exo motions, suggesting that the rendered spatial memory and pose controls provide complementary guidance.

\vspace{-0.5em}
\subsection{Ablation Study}
\label{sec:exp-ablation}

We ablate the conditioning signals and modules in \coolname{} and report the results in \cref{tab:quant-ablation} and \cref{fig:qual-ablate}. Starting from a Nymeria-fine-tuned VACE model that is conditioned only on the text prompt and one context image $\mathbf{x}_m$ (Base), we incrementally add point-cloud renderings (Point), per-point VAE appearance features (Feat), exo skeleton controls (Exo), drawn ego pose overlays (Drawn Ego), and our encoded ego pose tokens (Encoded Ego). Each variant is trained and evaluated independently under the same setup as the full model.

Point cloud conditioning provides the most significant improvement in visual fidelity, camera control, and semantic object consistency, showing that explicit geometry is crucial for the video generator to be 3D-consistent and to follow rapid head motion. Adding per-point appearance features further improves LPIPS/PSNR/SSIM, reduces perceptual discrepancy, and enhances texture sharpness (cf., \cref{fig:qual-ablate} top row). Exo and drawn ego overlays increase exo and ego human control metrics, respectively. Finally, encoded ego tokens yield the strongest gains for ego human control, improving Hand-F1 from 36.03 to 41.44 and Hand-mIoU from 19.36 to 24.76, indicating more reliable hand presence and segmentation overlap.

\begin{figure}[t!]
    \centering
    \setlength{\fboxsep}{0pt}
    \setlength{\fboxrule}{0.8pt}
    \newcommand{\smallimgwidth}{0.125\textwidth}
    \newcommand{\largeimgwidth}{0.125\textwidth}
    \scriptsize

    \resizebox{\linewidth}{!}{%
    {\renewcommand{\arraystretch}{1.1}
    \begin{tabular}{@{}c @{\hspace{0.2mm}} c @{\hspace{0.7mm}} c @{\hspace{0.2mm}} c @{\hspace{0.5mm}} | @{\hspace{0.5mm}} c @{\hspace{0.2mm}} c @{\hspace{0.7mm}} c @{\hspace{0.2mm}} c@{}}

    \multicolumn{4}{c@{\hspace{0.5mm}} | @{\hspace{0.5mm}}}{\textbf{Object Edits}} &
    \multicolumn{4}{c}{\textbf{Exo People Edits}} \\

    \multicolumn{2}{c}{Original} &
    \multicolumn{2}{c@{\hspace{0.5mm}} | @{\hspace{0.5mm}}}{Edited} &
    \multicolumn{2}{c}{Original} &
    \multicolumn{2}{c}{Edited} \\

    \fcolorbox{condcolor}{white}{\includegraphics[width=\smallimgwidth]{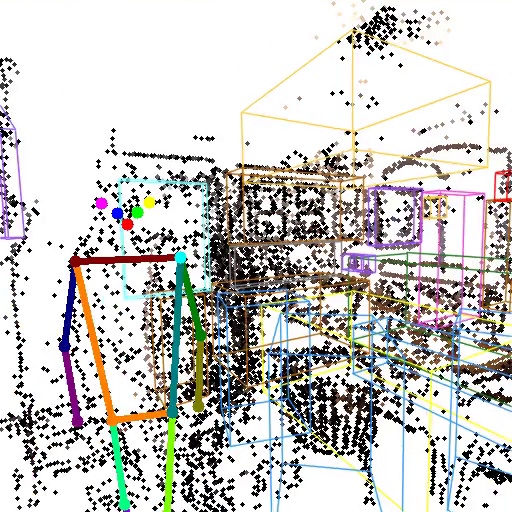}} &
    \fcolorbox{predcolor}{white}{\includegraphics[width=\largeimgwidth]{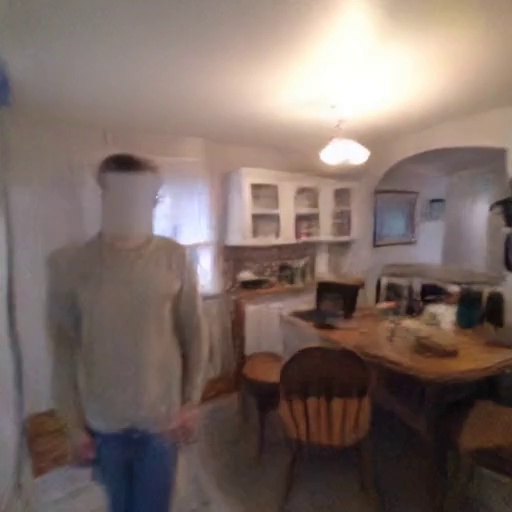}} &
    \fcolorbox{condcolor}{white}{\includegraphics[width=\smallimgwidth]{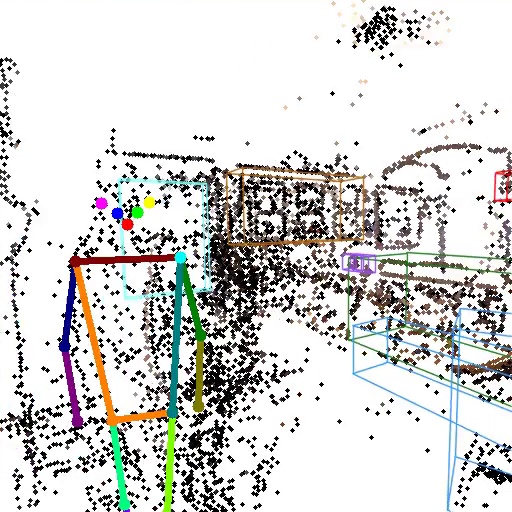}} &
    \fcolorbox{predcolor}{white}{\includegraphics[width=\largeimgwidth]{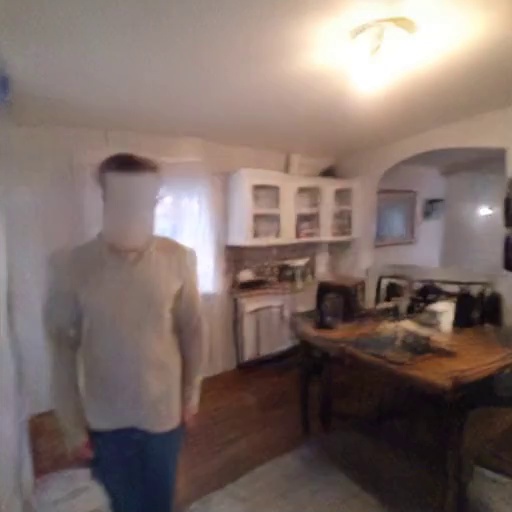}} &
    \fcolorbox{condcolor}{white}{\includegraphics[width=\smallimgwidth]{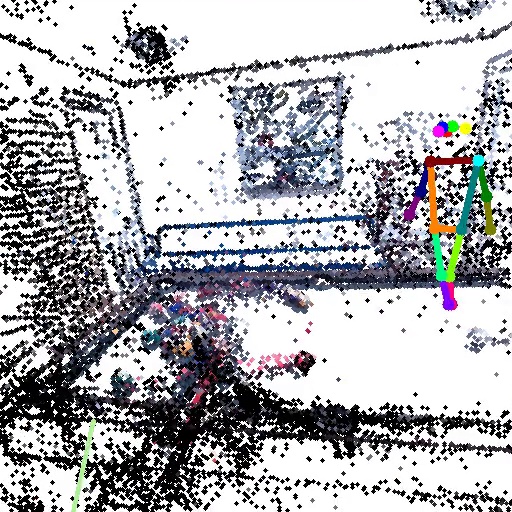}} &
    \fcolorbox{predcolor}{white}{\includegraphics[width=\largeimgwidth]{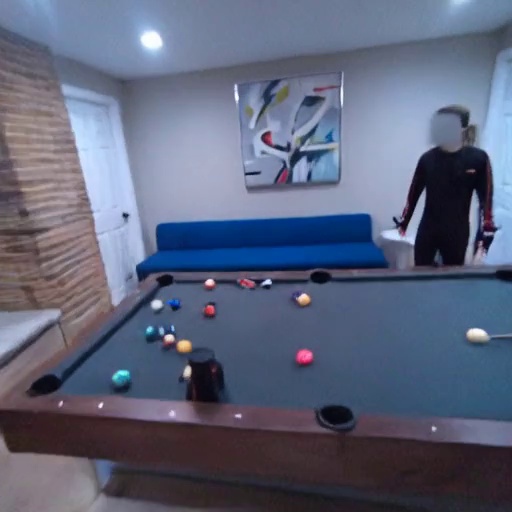}} &
    \fcolorbox{condcolor}{white}{\includegraphics[width=\smallimgwidth]{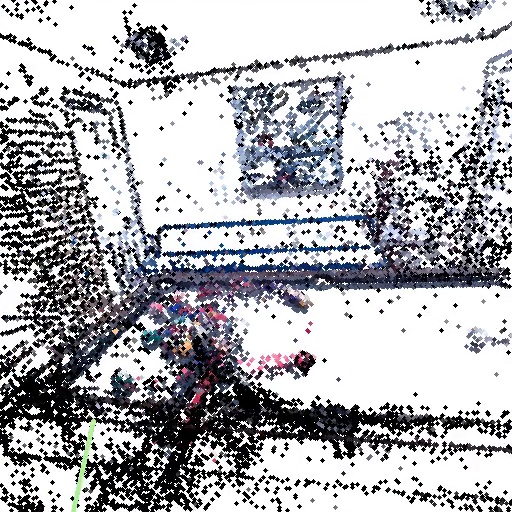}} &
    \fcolorbox{predcolor}{white}{\includegraphics[width=\largeimgwidth]{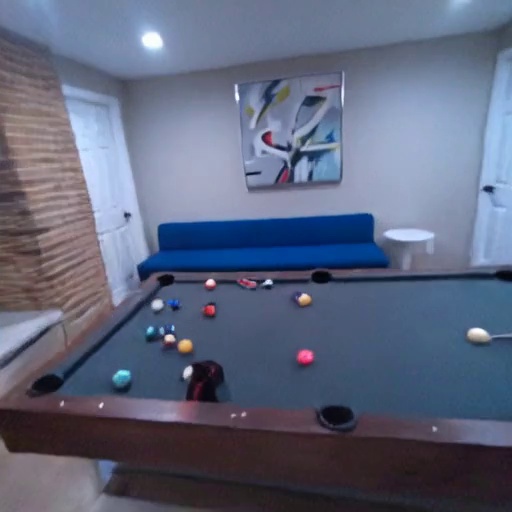}} \\

    \fcolorbox{condcolor}{white}{\includegraphics[width=\smallimgwidth]{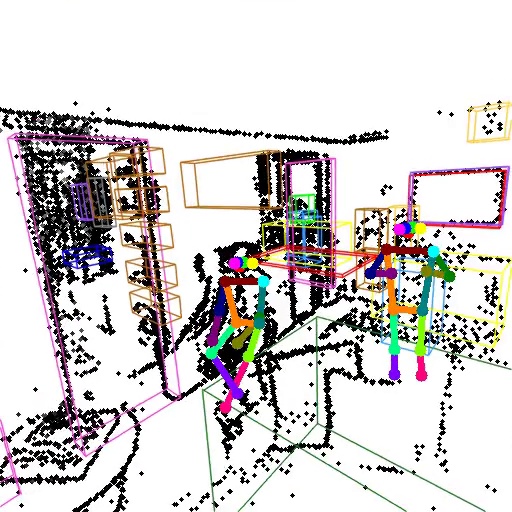}} &
    \fcolorbox{predcolor}{white}{\includegraphics[width=\largeimgwidth]{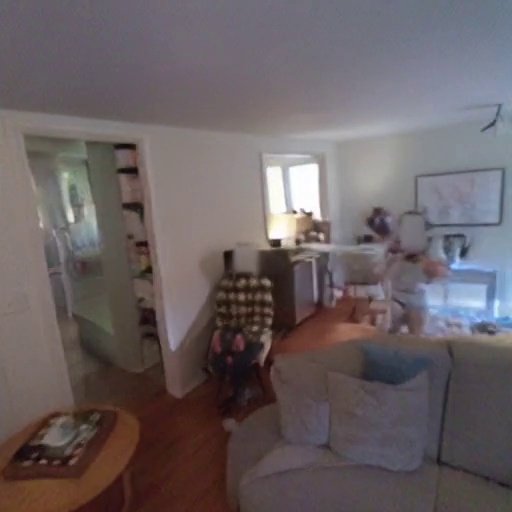}} &
    \fcolorbox{condcolor}{white}{\includegraphics[width=\smallimgwidth]{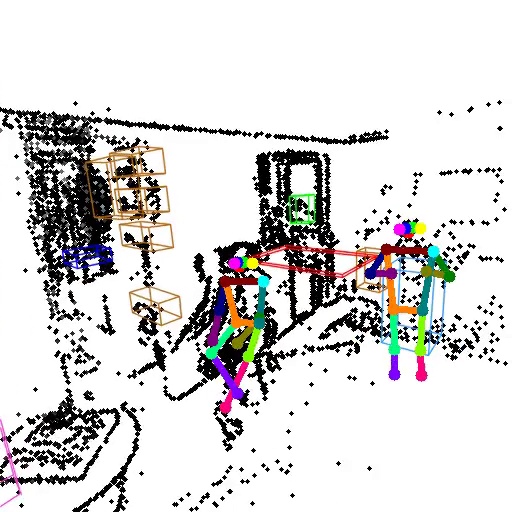}} &
    \fcolorbox{predcolor}{white}{\includegraphics[width=\largeimgwidth]{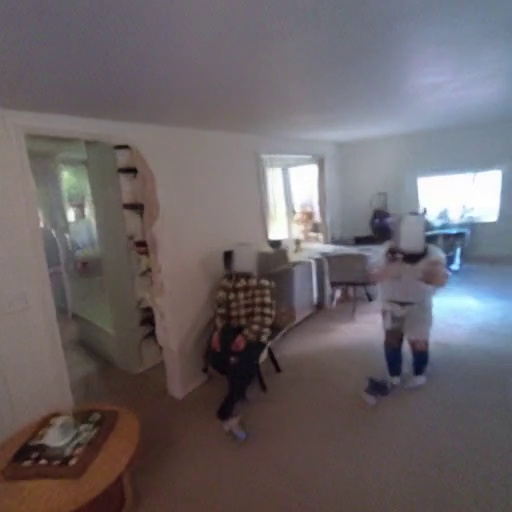}} &
    \fcolorbox{condcolor}{white}{\includegraphics[width=\smallimgwidth]{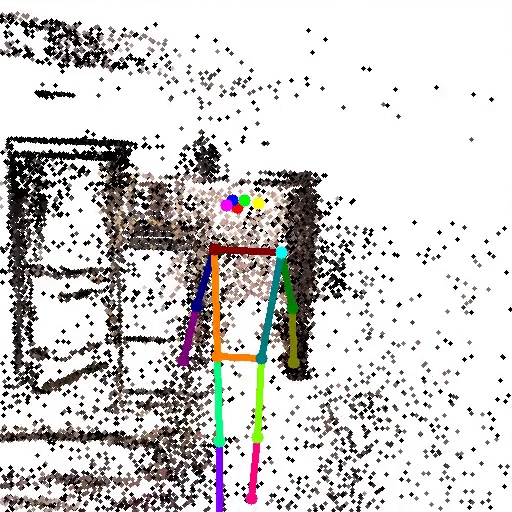}} &
    \fcolorbox{predcolor}{white}{\includegraphics[width=\largeimgwidth]{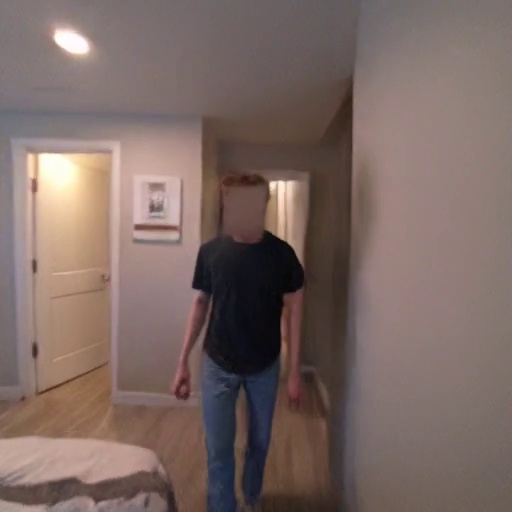}} &
    \fcolorbox{condcolor}{white}{\includegraphics[width=\smallimgwidth]{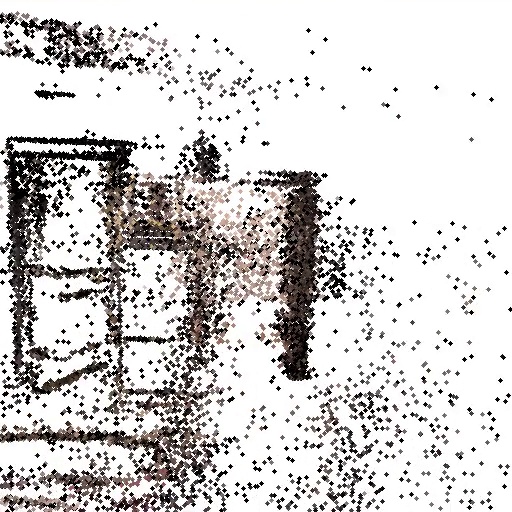}} &
    \fcolorbox{predcolor}{white}{\includegraphics[width=\largeimgwidth]{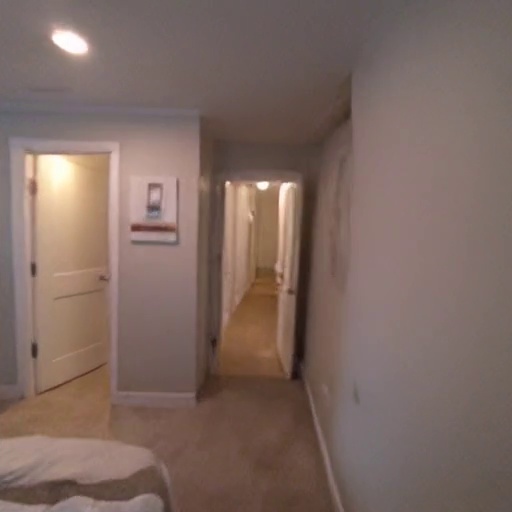}} \\

    \end{tabular}
    }%
    }%

    \vspace{-1em}
    \caption{Editing by manipulating explicit 3D memory and pose conditions. Left four columns: object removal by deleting the object's points from the 3D memory, which consistently removes it from generated frames. Right four columns: exo-person removal by removing the corresponding pose control. Both \textcolor{condcolor}{conditioning} and \textcolor{predcolor}{generated} frames before and after the edits are shown. See full videos on the project page.\vspace{-1.5em}}
    \label{fig:qual-scene-edit}
\end{figure}

\vspace{-0.5em}
\subsection{Scene Editing}

The explicit conditioning also enables intuitive edits. \cref{fig:qual-scene-edit} shows examples of removing objects by deleting points from the 3D memory and removing exo people by modifying the exo pose inputs. Because the edits are applied at the 3D representation and conditioning level, the resulting changes remain temporally and spatially consistent throughout the generated sequence. This behavior further illustrates the advantages of explicit 3D memory and pose conditioning for video generation. Additional results and videos are provided on the project page.

\vspace{-0.5em}
\section{Conclusion}
\label{sec:conclusion}

We presented \coolname{}, a controllable egocentric video generation framework that integrates 3D environmental memory with joint ego--exo human pose
control. A semi-dense point cloud-based spatial memory constructed from past views and augmented with per-point appearance features from video latents is rendered into target viewpoints to preserve scene geometry under strong ego motion. Exo humans are controlled via 2D skeleton renderings, while the ego wearer is conditioned by 3D body and wrist poses plus encoded ego pose tokens to robustly handle invisible body parts.
Experiments on Nymeria show that \coolname{} outperforms 3D-aware and pose-conditioned baselines on visual quality, camera-motion accuracy, object consistency, and ego and exo motion adherence.

\textbf{Limitations:}
While the spatial memory used in \coolname{} enables 3D consistency and camera control, it assumes a mostly static environment and models dynamics primarily through human motion, limiting performance when other objects move or the scene changes over long horizons.
\coolname{} provides motion control for ego and exo humans, but it does not explicitly condition on appearance. As a result, a person's appearance may drift when they leave the field of view and re-enter.

\section*{Acknowledgments}

The authors thank the Project Aria team for providing open-source support. The authors also thank Armen Avetisyan, Chris Sweeney, Fan Zhang, Jakob Engel, Nan Yang, Pierre Moulon, Ruocheng Wang, Tianwei Shen, Zhao Dong, and Zizhang Li for valuable and insightful discussions.

\bibliographystyle{splncs04}
\bibliography{main}

\clearpage
\section*{Appendix}
\appendix

\section{Qualitative Video Results}

On the \href{https://e3c-videogen.github.io}{project webpage}, we present multiple qualitative results. These include:
\begin{itemize}
    \item Video results from our full method, \coolname{}, including visualizations of the input conditions, generated videos, and ground-truth (GT) videos.
    \item Video results from ablation studies on per-point appearance features, exo human motion control, and ego human motion control.
    \item Video results from comparisons with baselines, including zero-shot and fine-tuned VACE \cite{jiang2025vace}, Splatfacto \cite{splatfacto2023}, Gen3C \cite{ren2025gen3c}, and VMem \cite{li2025vmem}.
    \item Video results demonstrating scene editing, including object removal, exo people removal, target camera changes, and motion and scene composition.
\end{itemize}

Please refer to the project webpage for qualitative results and visualizations. In the remainder of this appendix, we provide additional details on the method and experiments.

\section{Method Details}

In the following, we elaborate on implementation details.

\subsection{3D Spatial Memory Construction and Rendering}

In our implementation, we build the RGB point cloud and the feature point cloud in the spatial memory separately. The two memories use different voxel sizes in voxel pooling, and they are rendered into conditioning videos of different resolutions. Their construction and rendering processes are described below.

\subsubsection{RGB point cloud memory.}

As described in \cref{sec:method-memory}, we build the persistent 3D memory from the context frames $\mathbf{x}_{1:m}$ and render it onto the target camera views $\mathbf{T}_{1:n}$.
To build the 3D memory, we use the semi-dense points and observations (\texttt{semidense\_points.csv.gz} and \texttt{semidense\_observations.csv.gz}) produced by Project Aria MPS~\cite{aria_mps}. These files jointly provide a 3D point cloud in the world coordinate system and the corresponding observations at each capture timestamp.
From the MPS results, we extract the 3D points observed in the context frames $\pcd^{raw} \in \mathbb{R} ^{m\times N_{raw} \times 3}$ (the point cloud from each frame is padded to the same size $N_{raw}$ with NaN values).
Note that MPS runs on the grayscale SLAM cameras on Aria glasses, which have a wider field of view than the RGB camera.
In our experiments, to ensure fair comparison between baselines that only use RGB inputs, we only keep the points that fall in the RGB camera frustum.

From $\pcd^{raw}$, we first aggregate the points over time, remove all NaN entries, and merge duplicate 3D locations.
Each point is assigned a color from the context RGB images; if the same 3D point is observed multiple times, the implementation pools its colors across observations before aggregation.
After aggregation, the point set is voxel-downsampled using a voxel size of 0.01 m to obtain a compact memory, which corresponds to $\pcd$ and $\pcdcolor$ in \cref{sec:method-memory}.
This memory is then rendered into each target view by transforming the world points into the target camera frame, projecting them to the image plane, discarding invalid or out-of-bounds projections, and splatting them as pixels or small disks on the rendered image. The renderer uses the pooled point colors.

\subsubsection{Feature point cloud memory.}

In parallel with the RGB point memory, we build the feature point cloud $\pcdfeat$ in the 3D spatial memory. The context video is first converted to dense per-frame feature maps using the Wan2.1 video encoder \cite{wan2025wan}.
Note that in this encoding process, every context frame is treated as the first frame in a snippet, so there is no temporal downsampling, and the output feature maps have the same temporal resolution as the input video. The video encoder encodes each context frame to a 16-channel latent map at one-eighth spatial resolution.

The MPS point cloud $\pcd^{raw}$ is then projected into these feature maps, and point features are sampled by bilinear interpolation at the projected locations. The sampled descriptors are pooled in world space with a coarser voxel size of 0.02 m, which is larger than the RGB point-memory voxel size because feature descriptors are smoother and do not require the same geometric fidelity. The voxel feature memory is rendered back to the target camera by projecting voxel centers to the target feature-map grid and splatting the closest visible voxel at each pixel with a simple z-buffer rule.

\subsection{Feature Encoder}

The rendered feature maps are not passed through the video VAE encoder again. Instead, they are processed by the feature encoder, a separate lightweight 3D convolutional encoder.
The feature encoder follows a similar architecture to the Wan2.1 video encoder, but since its input is already at one-eighth spatial resolution, it does not need to further downsample the input.
Specifically, the feature encoder uses hidden width 1024, one residual block per stage, channel multipliers \([0.25, 0.125, 0.125]\), two temporal downsampling stages, and no additional spatial downsampling.
Its output is a 96-channel tensor with the same shape as the VACE encoder \cite{jiang2025vace}, so the rendered feature context can be added directly to the RGB VACE context before entering the diffusion transformer, as mentioned in \cref{sec:method-memory}.

The RGB conditioning video and the conditioning feature video play different roles. The RGB control stream starts from full-resolution video, so it uses the standard VACE encoder path \cite{jiang2025vace}. The feature stream starts from already-compressed, low-resolution multi-channel maps, so it uses a separate encoder with a different composition. This separation keeps the RGB pathway close to the pretrained video model while still allowing the rendered 3D feature memory to inject view-consistent appearance cues.

\subsection{Ego Pose Encoder}
\label{sec:appendix-egopose-tokenizer}

This subsection provides the network details for the persistent ego pose tokens introduced in \cref{sec:method-pose}. The encoder maps the target ego body joints \(\poseegoskel_{1:n}\) and wrist trajectories \(\poseegowrist_{1:n}\) to a compact sequence of motion-aware tokens that condition the video DiT through dedicated pose cross-attention.

\paragraph{Input and output.}
For a target clip of length \(n\) with \(J\) body joints, the inputs are
\begin{align}
\poseegoskel_{1:n} &\in \mathbb{R}^{n\times J\times 3}, \\
\poseegowrist_{1:n} &\in \mathbb{R}^{n\times 2\times 4\times 4}.
\end{align}
The encoder outputs a pose-token sequence
\begin{align}
\mathbf{H}^e \in \mathbb{R}^{(n+K)\times D_{\mathrm{DiT}}},
\end{align}
where \(K\) is the number of learned global summary tokens.

\paragraph{Per-joint motion features.}
Following the main text, the body branch starts from camera-frame 3D joints. For each joint trajectory, we compute first- and second-order temporal differences,
\begin{align}
\Delta \poseegoskel_{t,j} &= \poseegoskel_{t,j}-\poseegoskel_{t-1,j}, \\
\Delta^2 \poseegoskel_{t,j} &= \Delta \poseegoskel_{t,j}-\Delta \poseegoskel_{t-1,j},
\end{align}
with first-frame replication at the sequence boundary. Let \(h\) and \(p\) denote the head and pelvis joint indices. For frame \(t\) and joint \(j\), the encoder forms
\begin{align}
\mathbf{u}^{e}_{t,j}=\big[\poseegoskel_{t,j},\;\Delta\poseegoskel_{t,j},\;\Delta^2\poseegoskel_{t,j},\;\poseegoskel_{t,j}-\poseegoskel_{t,h},\;\poseegoskel_{t,j}-\poseegoskel_{t,p}\big] \in \mathbb{R}^{15}.
\end{align}
The first three terms preserve motion in the default camera coordinate frame, while the last two terms express each joint relative to the head and pelvis joints, providing complementary head-centered and pelvis-centered body context. Each \(\mathbf{u}^{e}_{t,j}\) is then linearly projected to a \(D_{\text{model}}\)-dimensional embedding, summed with a learnable joint-identity embedding, and passed through dropout.

\paragraph{Spatial mixing and frame token pooling.}
For each frame, the \(J\) joint embeddings are processed by \(n_{\text{spatial}}\) transformer encoder layers \cite{vaswani2017attention} that operate across the joint dimension and are shared over time. A learnable pooling query then applies multi-head attention (MHA) \cite{vaswani2017attention} over the full joint set, followed by layer normalization (LN) \cite{ba2016layer}, to produce one body token per frame:
\begin{align}
\mathbf{f}^{e}_{t}=\mathrm{LN}\Big(\mathrm{MHA}(\mathbf{q}^{e}_{\text{pool}},\mathbf{X}^{e}_{t},\mathbf{X}^{e}_{t})\Big) \in \mathbb{R}^{D_{\text{model}}}.
\end{align}

\paragraph{Wrist 6-DoF branch and gated fusion.}
For each wrist transform in \(\poseegowrist_{1:n}\), we extract translation \(\mathbf{p}\in\mathbb{R}^3\) and a continuous 6D rotation representation \(\mathbf{r}_{6\mathrm{D}}\in\mathbb{R}^{6}\) \cite{zhou2019continuity} given by the first two columns of the rotation matrix. The per-wrist feature vector is
\begin{align}
\mathbf{u}^{w}_{t,\ell} = [\mathbf{p}_{t,\ell},\Delta\mathbf{p}_{t,\ell},\Delta^2\mathbf{p}_{t,\ell},\mathbf{r}_{6\mathrm{D},t,\ell},\Delta\mathbf{r}_{6\mathrm{D},t,\ell},\Delta^2\mathbf{r}_{6\mathrm{D},t,\ell}] \in \mathbb{R}^{27},
\end{align}
where \(\ell\in\{L,R\}\) indexes the left and right wrists. These two wrist tokens are linearly embedded, augmented with wrist-identity embeddings, and pooled by a second learnable attention query in the style of set attention pooling \cite{lee2019settransformer} to produce a wrist token \(\mathbf{f}^{w}_{t}\in\mathbb{R}^{D_{\text{model}}}\). Body and wrist streams are fused with a learned gate \cite{srivastava2015highway, arevalo2017gated}:
\begin{align}
\mathbf{g}_{t} &= \sigma\big(W_g[\mathbf{f}^{e}_{t};\mathbf{f}^{w}_{t}]\big), \\
\tilde{\mathbf{f}}_{t} &= \mathbf{f}^{e}_{t} + \mathbf{g}_{t}\odot\mathbf{f}^{w}_{t}.
\end{align}
This gated residual form lets the encoder modulate how much wrist information contributes to each frame token.

\paragraph{Temporal modeling and global summary tokens.}
The fused frame tokens are augmented with sinusoidal temporal embeddings and processed by an \(n_{\text{temporal}}\)-layer temporal transformer \cite{vaswani2017attention} over \(t\in\{1,\dots,n\}\). Next, \(K\) learnable global queries attend to the full sequence of frame tokens, following the object-query style used in DETR \cite{carion2020detr}, to extract global summary tokens \(\mathbf{G}^e\in\mathbb{R}^{K\times D_{\text{model}}}\). The final output is
\begin{align}
\bar{\mathbf{H}}^e &= [\tilde{\mathbf{f}}_{1:n};\mathbf{G}^e] \in \mathbb{R}^{(n+K)\times D_{\text{model}}}, \\
\mathbf{H}^e &= \mathrm{Dropout}(W_{\text{dit}}\bar{\mathbf{H}}^e) + \mathbf{e}_{\text{pose}}.
\end{align}
We use \(D_{\text{model}}=512\), \(D_{\mathrm{DiT}}=1536\), \(K=4\), \(n_{\text{spatial}}=2\), and \(n_{\text{temporal}}=2\).

\begin{algorithm}[t]
\caption{Forward pass of the ego pose encoder}
\label{alg:egopose-tokenizer}
\begin{algorithmic}[1]
\Require Body joints \(\poseegoskel_{1:n}\in\mathbb{R}^{n\times J\times 3}\), wrist poses \(\poseegowrist_{1:n}\in\mathbb{R}^{n\times 2\times 4\times 4}\)
\Require Head index \(h\), pelvis index \(p\)
\Ensure Pose tokens \(\mathbf{H}^e\in\mathbb{R}^{(n+K)\times D_{\mathrm{DiT}}}\)
\State Compute \(\Delta\poseegoskel\) and \(\Delta^2\poseegoskel\) with first-frame replication
\State Build per-joint features \(\mathbf{u}^{e}=[\poseegoskel,\Delta\poseegoskel,\Delta^2\poseegoskel,\poseegoskel-\poseegoskel_{:,h},\poseegoskel-\poseegoskel_{:,p}]\)
\State Embed joints: \(\mathbf{X}^{e}\leftarrow\mathrm{Linear}(\mathbf{u}^{e}) + \mathrm{JointIDEmbed}\)
\State Apply spatial transformer stack over the joint dimension
\State Frame pooling: \(\mathbf{f}^{e}_{1:n}\leftarrow\mathrm{AttnPool}(\mathbf{X}^{e})\)
\State Extract per-wrist \(27\)-D features from translation, 6D rotation, and temporal derivatives
\State Wrist embedding + wrist-ID + wrist attention pooling to get \(\mathbf{f}^{w}_{1:n}\)
\State Compute gate \(\mathbf{g}_{1:n}\leftarrow\sigma(W_g[\mathbf{f}^{e}_{1:n};\mathbf{f}^{w}_{1:n}])\)
\State Fuse: \(\tilde{\mathbf{f}}_{1:n}\leftarrow\mathbf{f}^{e}_{1:n} + \mathbf{g}_{1:n}\odot\mathbf{f}^{w}_{1:n}\)
\State Add sinusoidal temporal embeddings to \(\tilde{\mathbf{f}}_{1:n}\)
\State Apply temporal transformer stack over time
\State Global extraction: \(\mathbf{Q}_{\text{global}}\) cross-attends to \(\tilde{\mathbf{f}}_{1:n}\) to produce \(K\) global tokens \(\mathbf{G}^{e}\)
\State Concatenate \(\bar{\mathbf{H}}^{e}\leftarrow[\tilde{\mathbf{f}}_{1:n};\mathbf{G}^{e}]\)
\State Project \(\bar{\mathbf{H}}^{e}\) to \(D_{\mathrm{DiT}}\), add pose embedding, apply output dropout
\State \Return \(\mathbf{H}^{e}\)
\end{algorithmic}
\end{algorithm}

\paragraph{Practical note.}
Compared with using only rasterized ego pose drawings in \(\condvid\), these persistent tokens preserve camera-frame 3D joint kinematics and wrist orientation cues in token space, which helps maintain motion control under self-occlusion, truncation, and out-of-view body motion.

\section{Experiment Details}

\subsection{Baselines}

\subsubsection{Gen3C \cite{ren2025gen3c}.}
We use the official Gen3C codebase and the released pretrained checkpoint.
Following the official setup, we use VIPE \cite{huang2025vipe} to obtain dense depth maps for each input context frame, where the camera extrinsics and intrinsics from MPS \cite{aria_mps} are used as the initialization for VIPE optimization. Gen3C warps the context frames into the target camera views according to the estimated dense depth maps and uses the warped images as conditioning for video generation.

The official inference script generates a 121-frame video chunk at a resolution of $1280\times704$ in each inference pass.
To align with this setup, we pad our 80-frame target camera trajectory to 121 frames by repeating the last camera poses.
We also resize the input context video and target camera intrinsics to $1280\times704$ before inference and resize the generated video back to $512\times512$ for evaluation and visualization.
Empirically, we find that this reduces the domain gap between the training data of Gen3C and our evaluation setting, leading to better results for Gen3C.

We adapt the ``video generation from multiview images'' example script provided by the authors.
The script selects only two context frames that jointly have the maximum overlap with the entire target camera trajectory and uses them to condition video generation. We find that this \textit{per-trajectory} selection provides insufficient view coverage for large view changes in our evaluation. Therefore, we modify the script to select two context frames for each target viewpoint independently (\textit{per-frame} selection), which significantly improves the performance of Gen3C in our setting.

\subsubsection{VMem \cite{li2025vmem}.}
We use the official VMem codebase and the released pretrained checkpoint.
The released VMem model takes 4 context frames as input and generates 4 target frames in each inference pass. Therefore, we generate the full 80-frame video autoregressively using 20 inference passes.

In each inference pass, 4 context frames are selected from VMem's surfel-indexed memory and used to condition the video generator. We conduct experiments ablating whether the generated frames from each inference pass are inserted back into the memory for future generation, and how the camera poses are normalized at each inference pass. However, we find that these factors minimally affect VMem's performance in our setting.

The input is resized to $576\times576$ for model inference, and the generated video is resized back to $512\times512$ for evaluation and visualization.

\subsubsection{Splatfacto \cite{splatfacto2023}.}
We use the 3D Gaussian splatting-based \cite{kerbl2023gaussiansplatting} method Splatfacto provided by NerfStudio \cite{splatfacto2023}. In our experiments, Splatfacto uses the camera poses and intrinsics estimated by MPS \cite{aria_mps} and initializes the 3D Gaussians using the MPS semi-dense point cloud.
One 3D Gaussian splatting model is trained and evaluated individually on each snippet. We use the default training settings provided by NerfStudio with 30,000 training iterations.

\subsubsection{Comparison with PEVA \cite{bai2025peva} and EgoControl \cite{pallotta2025egocontrol}.}
Since PEVA and EgoControl have not released their code or model weights, we reproduce their evaluation setup to obtain the comparison results reported in \cref{tab:peva_comparison}.
Specifically, we obtain the full list of 300 snippets used in the PEVA evaluation from the authors.
As their data split differs from ours, we train a separate \coolname{} model on sequences from the Nymeria dataset that are not included in the PEVA evaluation set and evaluate it on the PEVA snippets.

To further align the evaluation protocol, for this separately trained model, we disable the devignetting and undistortion preprocessing and use the original fisheye images.
As mentioned in \cref{sec:exp-results}, this evaluation uses a 15-frame context window from the past 3.75 seconds and evaluates a single generated frame 2 seconds into the future.

Following the suggestion of the PEVA authors, we use the evaluation script from NWM\footnote{\href{https://github.com/facebookresearch/nwm/blob/main/isolated\_nwm\_eval.py}{https://github.com/facebookresearch/nwm/blob/main/isolated\_nwm\_eval.py}} \cite{bar2025nwm} to compute the LPIPS, DreamSim, and FID metrics reported in \cref{tab:peva_comparison}.

\subsection{Evaluation Metrics}

\paragraph{Notation.}
Let the GT target video be
\[
\mathbf{y}_{1:n}\in\mathbb{R}^{n\times \imgh\times \imgw\times 3},
\]
and let the generated video be
\[
\hat{\mathbf{y}}_{1:n}\in\mathbb{R}^{n\times \imgh\times \imgw\times 3}.
\]
When spatial resolutions differ, generated frames are resized to the GT resolution before comparison. Let \(\mathbf{T}_{1:n}=\{T_t\}_{t=1}^n\), \(T_t=(R_t,\mathbf{t}_t)\in SE(3)\), denote the GT camera trajectory, and let \(\poseexo_{1:n}\) denote the GT exocentric 2D human poses.

\subsubsection{Object Consistency (Obj-F1, Obj-mIoU).}
We run OWLv2~\cite{minderer2023owlv2} on each aligned frame of \(\mathbf{y}_{1:n}\) and \(\hat{\mathbf{y}}_{1:n}\), using the default indoor prompt set and a score threshold of \(0.2\). This yields detection sets
\[
\mathcal{D}^{\text{obj}}_t,\ \widehat{\mathcal{D}}^{\text{obj}}_t.
\]
Each detection contains a bounding box, and GT/generated boxes are matched independently in each frame by Hungarian assignment on cost \(1-\operatorname{IoU}\), keeping only matches with IoU at least \(\tau_{\text{obj}}=0.5\). Obj-F1 is then computed from the aggregate TP/FP/FN counts over all aligned frames. Obj-mIoU is the mean of frame-wise matched-box IoU scores,
\[
m_t=
\begin{cases}
\frac{1}{|\mathcal{M}_t|}\sum_{(i,j)\in\mathcal{M}_t}\operatorname{IoU}(\mathbf{b}_i,\hat{\mathbf{b}}_j), & |\mathcal{M}_t|>0\\
0, & |\mathcal{M}_t|=0,\ |\mathcal{D}^{\text{obj}}_t|+|\widehat{\mathcal{D}}^{\text{obj}}_t|>0\\
1, & |\mathcal{D}^{\text{obj}}_t|=|\widehat{\mathcal{D}}^{\text{obj}}_t|=0
\end{cases},
\]
where \(\mathcal{M}_t\) is the filtered match set at frame \(t\). The reported score is
\[
\text{Obj-mIoU}=\frac{1}{n}\sum_{t=1}^{n} m_t.
\]

\subsubsection{Camera Motion Control (TErr, RErr).}
For camera control, we estimate camera poses from generated frames \(\hat{\mathbf{y}}_{1:n}\) using VIPE~\cite{huang2025vipe}, yielding \(\widehat{T}_t=(\widehat{R}_t,\widehat{\mathbf{t}}_t)\). We then align \(\widehat{T}_{1:n}\) to the GT trajectory \(\mathbf{T}_{1:n}\) using a global Sim(3) transform
\[
S=(s,R_S,\mathbf{t}_S),\qquad
(\widehat{R}'_t,\widehat{\mathbf{t}}'_t)=\big(R_S\widehat{R}_t,\ sR_S\widehat{\mathbf{t}}_t+\mathbf{t}_S\big),
\]
with \(S\) fit once over the whole trajectory. Per-frame translation and rotation errors are
\[
e^{\text{trans}}_t=\|\widehat{\mathbf{t}}'_t-\mathbf{t}_t\|_2,
\]
\[
e^{\text{rot}}_t=\frac{180}{\pi}\arccos\!\left(
\operatorname{clip}\!\left(\frac{\operatorname{tr}\!\left((\widehat{R}'_t)^\top R_t\right)-1}{2},-1,1\right)
\right).
\]
The reported metrics are frame averages,
\[
\mathrm{TErr}=\frac{1}{n}\sum_{t=1}^{n} e^{\text{trans}}_t,
\qquad
\mathrm{RErr}=\frac{1}{n}\sum_{t=1}^{n} e^{\text{rot}}_t.
\]
In our table, \(\mathrm{TErr}\) is displayed in centimeters.

\subsubsection{Ego Hand Control (Hand-F1, Hand-mIoU).}
We run SAM3~\cite{carion2025sam3} on each aligned frame of \(\mathbf{y}_{1:n}\) and \(\hat{\mathbf{y}}_{1:n}\). Since the ego human in the Nymeria dataset always wears a mocap suit that covers part of their hands and wrists, we use the text prompts ``hand'' and ``glove'' for SAM3. We use a confidence threshold of \(0.3\), and masks smaller than 1536 pixels are filtered out to remove detected hands on exo people. This produces binary hand masks
\[
\mathbf{M}^{\text{hand}}_t,\ \widehat{\mathbf{M}}^{\text{hand}}_t \in \{0,1\}^{\imgh\times\imgw}
\]
for the GT and generated frames, respectively. Hand presence indicators are
\[
z_t=\mathbbm{1}\!\left[\sum \mathbf{M}^{\text{hand}}_t>0\right],\quad
\hat{z}_t=\mathbbm{1}\!\left[\sum \widehat{\mathbf{M}}^{\text{hand}}_t>0\right].
\]
Hand-F1 is computed from the aggregate TP/FP/FN counts over \((z_t,\hat{z}_t)\) across all aligned frames. For mask overlap, the per-frame hand IoU is
\[
\mathrm{IoU}^{\text{hand}}_t=
\begin{cases}
\frac{|\mathbf{M}^{\text{hand}}_t\cap \widehat{\mathbf{M}}^{\text{hand}}_t|}
{|\mathbf{M}^{\text{hand}}_t\cup \widehat{\mathbf{M}}^{\text{hand}}_t|}, & |\mathbf{M}^{\text{hand}}_t\cup \widehat{\mathbf{M}}^{\text{hand}}_t|>0\\
1, & \text{otherwise}
\end{cases}.
\]
The reported Hand-mIoU is the mean of \(\mathrm{IoU}^{\text{hand}}_t\) over all \(n\) aligned frames, so empty-empty frames contribute \(1\).

\subsubsection{Exo Human Gesture Control (Exo-F1, PCK@10\%).}
From \(\mathbf{y}_{1:n}\) and \(\hat{\mathbf{y}}_{1:n}\), we extract per-frame exo person boxes and keypoints using YOLO-X + ViTPose. The frames are resized to \(640\times 640\) for inference and the resulting boxes/keypoints are mapped back to the original image resolution. Detections whose average keypoint confidence is below \(0.5\) are discarded. This yields
\[
(\mathcal{B}_t,\mathcal{K}_t),\ (\widehat{\mathcal{B}}_t,\widehat{\mathcal{K}}_t).
\]
For each frame \(t\), we match \(\mathcal{B}_t\) and \(\widehat{\mathcal{B}}_t\) by Hungarian assignment on cost \(1-\operatorname{IoU}\), retaining matches with \(\operatorname{IoU}\ge\tau_{\text{exo}}\), where \(\tau_{\text{exo}}=0.2\). Exo-F1 is computed from the aggregate TP/FP/FN counts over all aligned frames.

Pose accuracy is computed only on matched person pairs. For a matched pair \(m\) and keypoint \(j\), let \(A_m\) be the GT box area, with GT keypoint \(\mathbf{k}_{m,j}\) and generated keypoint \(\widehat{\mathbf{k}}_{m,j}\). The normalized error is
\[
e_{m,j}=\frac{\|\widehat{\mathbf{k}}_{m,j}-\mathbf{k}_{m,j}\|_2}{\sqrt{A_m}}.
\]
The reported normalized MPJPE averages \(e_{m,j}\) over all valid matched keypoints. PCK@10\% is
\[
\mathrm{PCK@10\%}
=
\frac{1}{|\mathcal{V}|}\sum_{(m,j)\in\mathcal{V}} \mathbbm{1}[e_{m,j}<0.10].
\]
If there are no valid matched keypoints, the pose metrics are undefined and are not considered in the normalization. Therefore, as noted in the main text, PCK@10\% should be interpreted together with Exo-F1.

\end{document}